\documentclass[10pt,twocolumn,letterpaper]{article}

\usepackage{iccv}
\usepackage{times}
\usepackage{epsfig}
\usepackage{graphicx}
\usepackage{amsmath}
\usepackage{amssymb}
\usepackage{amsfonts}
\usepackage{subcaption}
\usepackage{tabularx}
\usepackage{bbm}
\usepackage{color}
\usepackage{mathtools}
\usepackage{multirow}
\usepackage[dvipsnames]{xcolor}
\usepackage{booktabs}
\usepackage{enumitem} %
\usepackage{float}
\makeatletter
\definecolor{darkolivegreen}{rgb}{0.5, 0.7, 0.3}

\newcommand{\KY}[1]{}

\newcommand{\PF}[1]{}

\newcommand{\PE}[1]{}
\newcommand{\et}[1]{#1}
\newcommand{\ET}[1]{}

\newcommand{\AM}[1]{}
\newcommand{\TODO}[1]{}

\newcommand{\bI}{\mathbf{I}}

\newcommand{\fig}[1]{Fig.~\ref{fig:#1}}

\newcommand{\tbl}[1]{Table~\ref{tbl:#1}}
\newcommand{\secref}[1]{Section~\ref{sec:#1}}
\newcommand{\refsec}[1]{Section~\ref{sec:#1}}

\newcommand{\loss}{\mathcal{L}}

\newcommand{\imsize}{1cm}
\newcommand{\hspacer}{0.3mm}
\newcommand{\vspacer}{0.1mm}
\newcommand{\parag}[1]{\vspace{-3mm}\paragraph{#1}}

\newcommand{\pa}{\mathbf{P}^a}
\newcommand{\pb}{\mathbf{P}^b}
\newcommand{\da}{\mathbf{f}^a}
\newcommand{\db}{\mathbf{f}^b}

\newcommand{\danc}{\mathbf{f}^A}
\newcommand{\dpos}{\mathbf{f}^+}
\newcommand{\dneg}{\mathbf{f}^-}

\newcommand{\scale}{\sigma}
\newcommand{\ori}{\theta}
\newcommand{\kp}{\mathbf{p}}
\newcommand{\mult}{\lambda}

\usepackage[pagebackref=true,breaklinks=true,letterpaper=true,colorlinks,bookmarks=false]{hyperref}
\usepackage{flushend}
\iccvfinalcopy %

\def\iccvPaperID{2277} %

\ificcvfinal\fi

\newcommand{\blfootnote}[1]{%
  \begingroup
  \renewcommand\thefootnote{}\footnote{#1}%
  \addtocounter{footnote}{-1}%
  \endgroup
}
\begin{document}

\title{Beyond Cartesian Representations for Local Descriptors}

\author{%
Patrick Ebel$^1$, Anastasiia Mishchuk$^1$, Kwang Moo Yi$^2$, Pascal Fua$^1$, Eduard Trulls$^3$\\{\small $^1$Computer Vision Lab, \'{E}cole Polytechnique F\'{e}d\'{e}rale de Lausanne} \\
{\small $^2$Visual Computing Group, University of Victoria {} {} {} $^3$Google Switzerland} \\
{\small{\tt \{firstname.lastname\}@epfl.ch}, {\tt kyi@uvic.ca}, {\tt trulls@google.com}}
}

\maketitle
\blfootnote{
  This research
  was partially funded
  by Google's Visual Positioning System, the Swiss National Science Foundation,
    the Natural Sciences and Engineering Research Council of Canada,
    and
    by Compute Canada.}

\begin{abstract}
The dominant approach for learning local patch descriptors relies on small image regions whose scale must be properly estimated a priori by a keypoint detector. In other words, if two patches are not in correspondence, their descriptors will not match. A strategy often used to alleviate this problem is to ``pool'' the pixel-wise features over log-polar regions, rather than regularly spaced ones.

By contrast, we propose to extract the ``support region'' directly with a log-polar sampling scheme. We show that this provides us with a better representation by simultaneously oversampling the immediate neighbourhood of the point and undersampling regions far away from it.
We demonstrate that this representation is particularly amenable to learning descriptors with deep networks. Our models can match descriptors across a much wider range of scales than was possible before, and also leverage much larger support regions without suffering from occlusions. We report state-of-the-art results on three different datasets.

\end{abstract}

\section{Introduction}
\label{sec:intro}

Keypoint matching has played a pivotal role in computer vision for well over a decade.
This is clearly demonstrated by the fact that SIFT~\cite{Lowe04} remains the most cited paper in computer vision history.
While many areas of computer vision are currently dominated by dense deep networks, that is, methods that take entire images as input, some problems remain best approached using sparse features.
For example, despite recent attempts at tackling 6DOF pose estimation using dense networks, 
the top-performing models for wide-baseline stereo and large-scale Structure-from-Motion (SfM) still rely on sparse features~\cite{Yi18a,Zheng18,Sattler18}.

As a result, the quest for ever-improving local feature descriptors goes on~\cite{Lowe04,Bay06,Winder07,Tola08,Simonyan14,Han15,Zagoruyko15,Simo-Serra15,Tian17,Mishchuk17,Wei18,Keller18,Luo18,He18b,Luo19,Dusmanu19,Revaud19}. 
These methods all seek to achieve invariance to small changes in location, orientation, scale, perspective, and illumination, along with imaging artefacts and partial occlusions. Most descriptors, however, whether learned or hand-crafted, operate on SIFT-like keypoints and thus rely on simple heuristics to estimate the scale. If the scales for two keypoints do not correspond, neither will the support regions used to extract their descriptors, which is widely accepted as an unrecoverable situation.
This is damaging because scale detection is often unreliable.

In this paper we demonstrate that this does not {\em need} to be the case. To this end, we go beyond the current paradigm for local descriptors,
which we call the {\em cartesian} approach.
This paradigm confines local descriptors to small, regularly sampled regions and relies on accurate scale estimates.
By contrast,
we posit that extracting the {\em support region} with a log-polar sampling scheme allows us to generate a better local representation by oversampling the immediate neighborhood of the point. We show that this approach is conducive to learning scale-invariant descriptors with off-the-shelf deep networks, enabling us to match keypoints across mismatched scales;
see \fig{fpr95bins}.
Furthermore, we demonstrate that this %
representation is far less sensitive to occlusions or background motion than its cartesian counterpart, which allows us to exploit much larger image regions than was possible before to further boost performance.

Note that while log-polar representations have been used extensively by local features, this has typically involved log-polar aggregation of local statistics that are still computed on the cartesian image grid. By contrast, we propose to {\em warp} the patch using a log-polar sampling scheme and learn an optimal descriptor on this data. \fig{sampling} illustrates the difference between these two approaches.

In short, we propose a new approach to represent local patches and show how to leverage it to achieve scale invariance.
In the remainder of the paper, we first briefly review how scale has been handled in the vast body of literature pertaining to matching descriptors, whether learned or designed. We then describe our method and show that it outperforms the state of the art on several challenging datasets.

\section{Related works}
\label{sec:related}

In this section we first review techniques representative of the many that have been proposed to achieve scale invariance for local feature matching, with and without explicit scale detection. Next, we discuss approaches to learning models for patch descriptors.
Finally, we study the use of log-polar representations in local features. For a thorough, up-to-date survey on local features please refer to~\cite{Csurka18}.

\vspace{-4mm}
\paragraph{Scale Invariance via Scale Detection.}

The vast majority of work in the literature assumes that scale estimation is handled by the keypoint detector and that keypoints can be put in correspondence only if their scales match.
This includes classical hand-crafted pipelines such as SIFT~\cite{Lowe04} or SURF~\cite{Bay06}.
Image measurements are then aggregated over a correspondingly-sized support region to extract the descriptor. As a result, errors in this {\it a priori} scale estimation cannot be recovered from, and the affected keypoints are simply written off as potential correspondences.
\parag{Two-stage pipelines.}
Special strategies can be used for rigid matching under large zoom.
Zhou \etal~\cite{Zhou17e} propose a two-stage approach to first coarsely register the image in scale-space and then narrow down the search scope to matches of commensurate scale.
Shan \etal~\cite{Shan14} assume that dense SfM models are available, along with an approximate pose,
and synthesize ground views from aerial viewpoints using the 3D model, for aerial-to-ground matching.
Both methods rely on SIFT features and would directly benefit from improved, scale-invariant descriptors such as ours.

\parag{Scale Invariance without Scale Detection.}

A simple way to achieve scale invariance is to concatenate multi-scale descriptors and find the best match among them. This was done in~\cite{Xompero18} to improve robustness against scale changes with ORB features~\cite{Rublee11}. Scale-Less SIFT (SLS)~\cite{Hassner12} goes beyond that and exploits the observation that SIFT descriptors do not change drastically over close, contiguous scales, which suggests that they are embedded in a low-dimensional space. This observation can be used to find a representation more compact than their concatenation. The resulting feature vectors are still high-dimensional (8k) but can be reduced by PCA to a 512-dimensional vector. However, this requires a singular value decomposition for each keypoint to find its subspace, which is very costly.

The Scale and rotation-Invariant Descriptor (SID)~\cite{Kokkinos12}
samples axis-aligned derivatives over a log-polar grid, along with incremental smoothing over image regions further away from the keypoint. Thus, scale changes and rotations result in translations on the measurement matrix. Using the Fourier Transform Modulus of this signal, which is translation-invariant, makes the descriptors scale- and rotation-invariant. However, SID requires fine sampling over large support regions,
which fails in real-world scenarios with viewpoint changes and occlusions. Seg-SID~\cite{Trulls15} addresses this shortcoming by leveraging segmentation cues to suppress image measurements from image regions not associated to the keypoint, but this requires image-level segmentation and is failure-prone. SID also suffers from high dimensionality ($\sim$3k).

More importantly, both SID and SLS were designed for dense matching with SIFT Flow~\cite{Liu08b} as a back-end and are not suitable for large-scale reconstruction due to their computational cost. Finally, they both rely on hand-crafted features and cannot compete with the machine learning models that currently dominate the field. We now turn to these.

\parag{Learned Descriptors.}

Early works applied PCA to SIFT~\cite{Ke04}, learned comparison metrics~\cite{Strecha12}, or learned descriptors with convex optimization~\cite{Simonyan14}.
Current research on patch descriptors is dominated by convolutional neural networks. MatchNet~\cite{Han15} and DeepCompare~\cite{Zagoruyko15} train descriptor extraction and distance metric networks using a Siamese architecture. DeepDesc~\cite{Simo-Serra15} uses hard positive and negative mining to learn discriminative features.
A triplet-based loss is introduced in~\cite{Balntas16b}. L2-Net~\cite{Tian17} improves the loss function by enforcing similarity in the intermediate feature maps and penalizing highly correlated descriptor bins. HardNet~\cite{Mishchuk17} extends the formulation of~\cite{Simo-Serra15} to mine over all the samples in the batch. In~\cite{He18b}, mining heuristics are replaced by a differentiable approximation of the average precision metric that is then used for optimization. Spectral pooling is introduced in~\cite{Wei18} to deal with geometric transformations. An alternative to siamese- and triplet-based loss functions is proposed in~\cite{Keller18} to address their shortcomings.
GeoDesc~\cite{Luo18} uses geometry constraints for optimization. ContextDesc~\cite{Luo19} incorporates global context, and geometric context from the keypoint distribution.

All of the deep methods, except~\cite{Luo18,Luo19}, are trained on the same dataset~\cite{Brown10}, which consists of patches pre-extracted on keypoints using Difference of Gaussians (DoG)~\cite{Lowe04} or multi-scale Harris corners~\cite{Harris88}. Only keypoints that survive a 3D reconstruction by Structure from Motion (SfM) are considered, and similarly to the traditional approach, the learned models are simply expected to fail if the detector fails first. To the best of our knowledge there is no learning-based method that explicitly addresses scale invariance.

Another line of works comprises those that use deep architectures to learn
keypoints and descriptors jointly.
LIFT~\cite{Yi16b} is trained on patches extracted around SIFT keypoints with corresponding scales, similarly to the previous methods. LF-Net~\cite{Ono18} learns to detect the scale with self-supervision, but in practice seems to perform best over a very narrow set of scales. SuperPoint~\cite{DeTone17b}
learns scale invariance at the descriptor level, which works for visual odometry but breaks in more generalized problems. D2-Net~\cite{Dusmanu19} focuses on difficult imaging conditions and relies on a single network for detection and description. R2D2~\cite{Revaud19} applies L2-Net convolutionally while penalizing repeatable but non-discriminative patches.

\parag{Leveraging Polar representations.}

\renewcommand{\arraystretch}{0.5}
\renewcommand{\tabcolsep}{0.3mm}
\renewcommand{\imsize}{1.9cm}
\begin{figure}
\centering
\begin{tabular}{@{}ccccc@{}}
\raisebox{2.5em}{{\footnotesize 1x}} &
\includegraphics[width=\imsize]{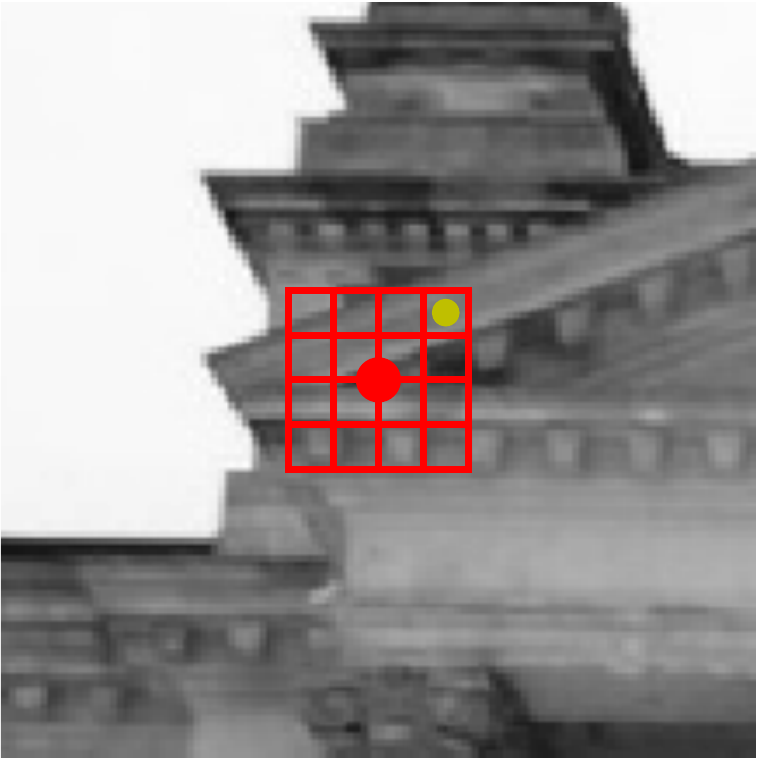} &
\includegraphics[width=\imsize]{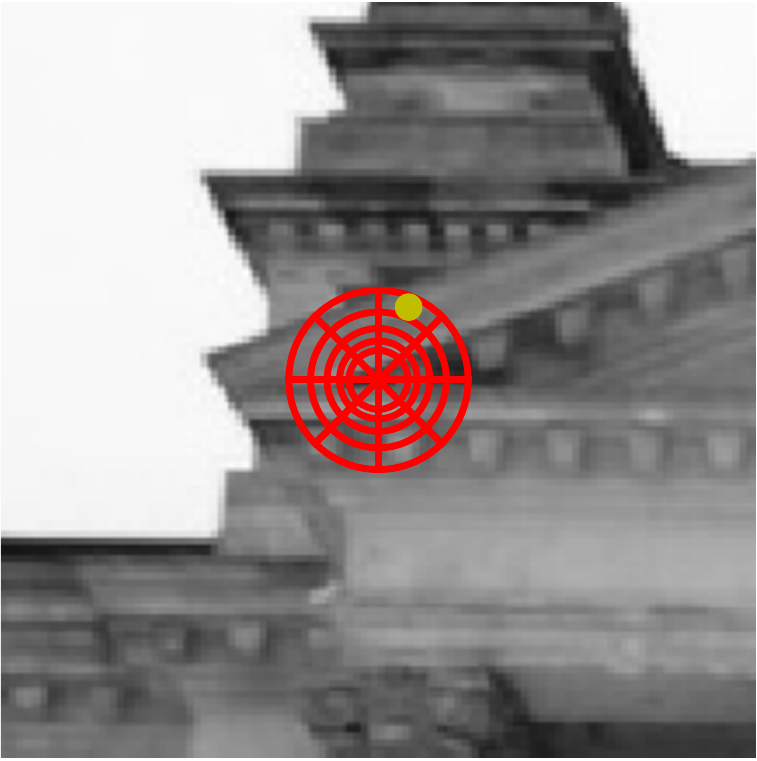} &
\includegraphics[width=\imsize]{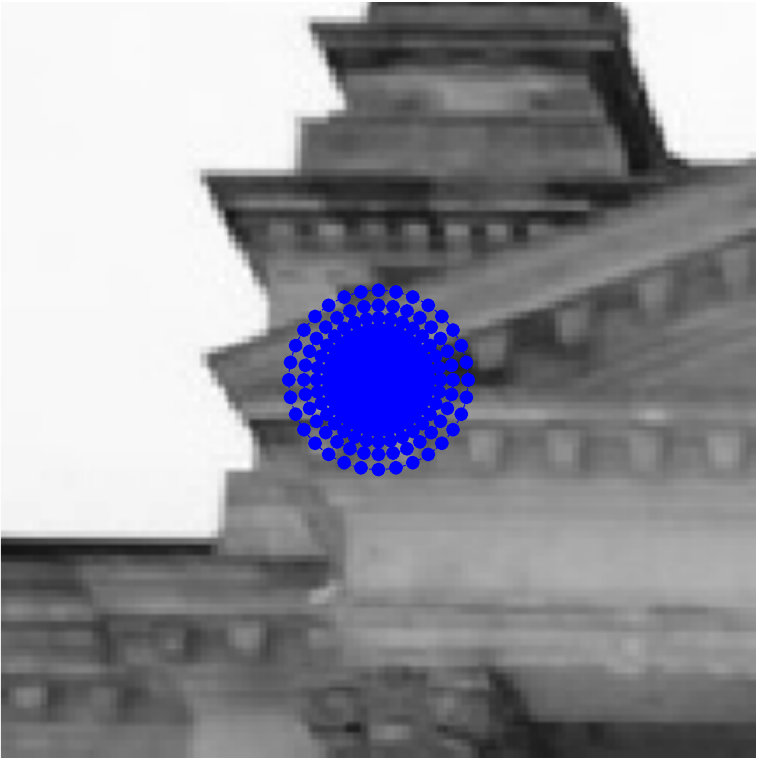} &
\includegraphics[width=\imsize]{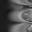} \\
\raisebox{2.5em}{{\footnotesize 2x}} &
\includegraphics[width=\imsize]{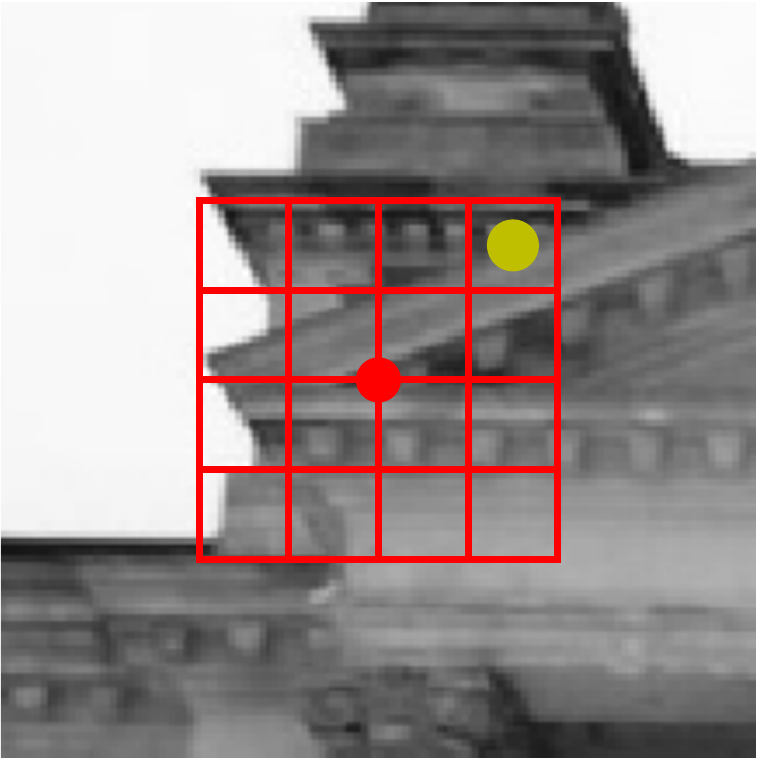} &
\includegraphics[width=\imsize]{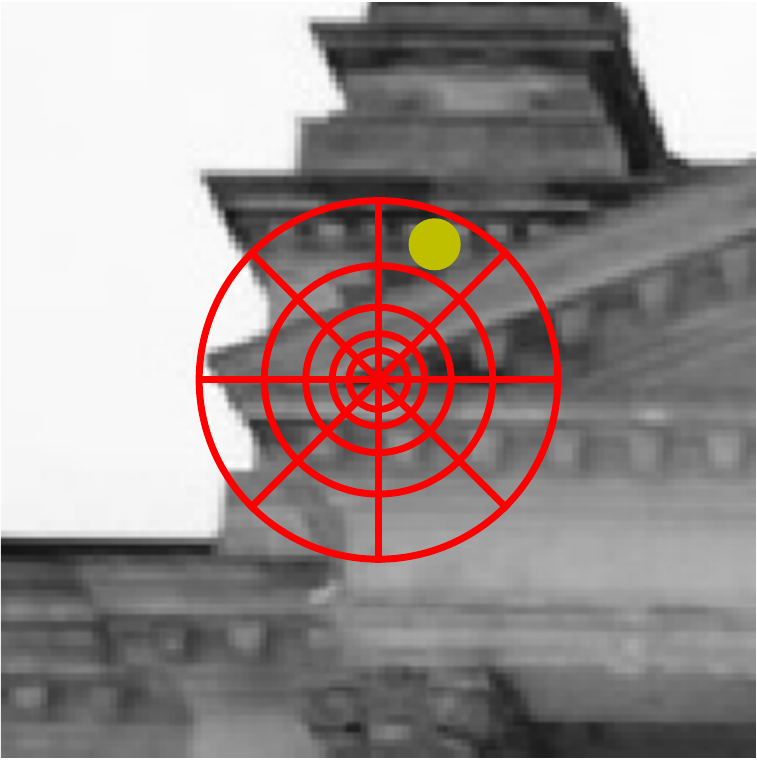} &
\includegraphics[width=\imsize]{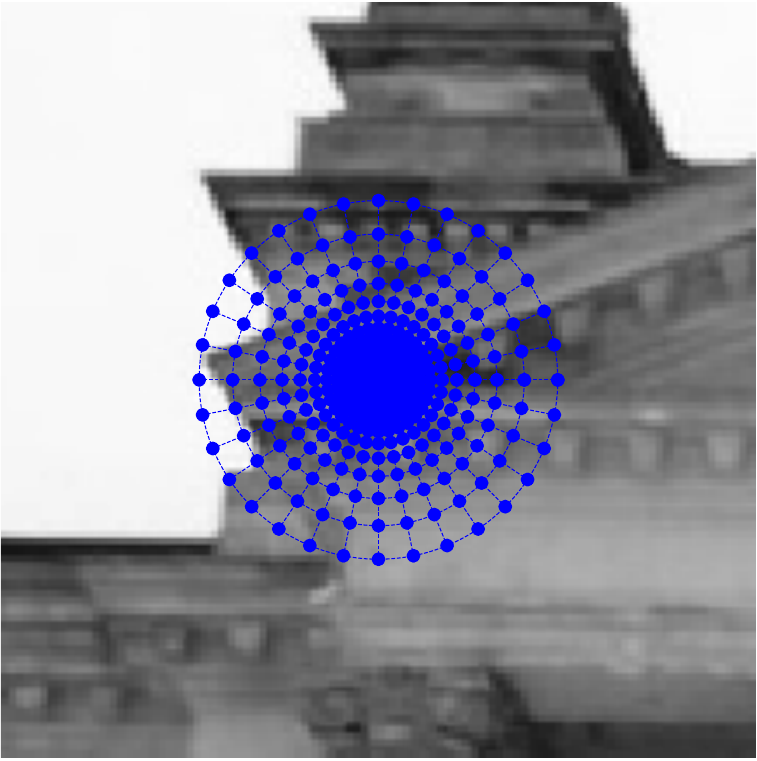} &
\includegraphics[width=\imsize]{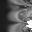} \\
\raisebox{2.5em}{{\footnotesize 4x}} &
\includegraphics[width=\imsize]{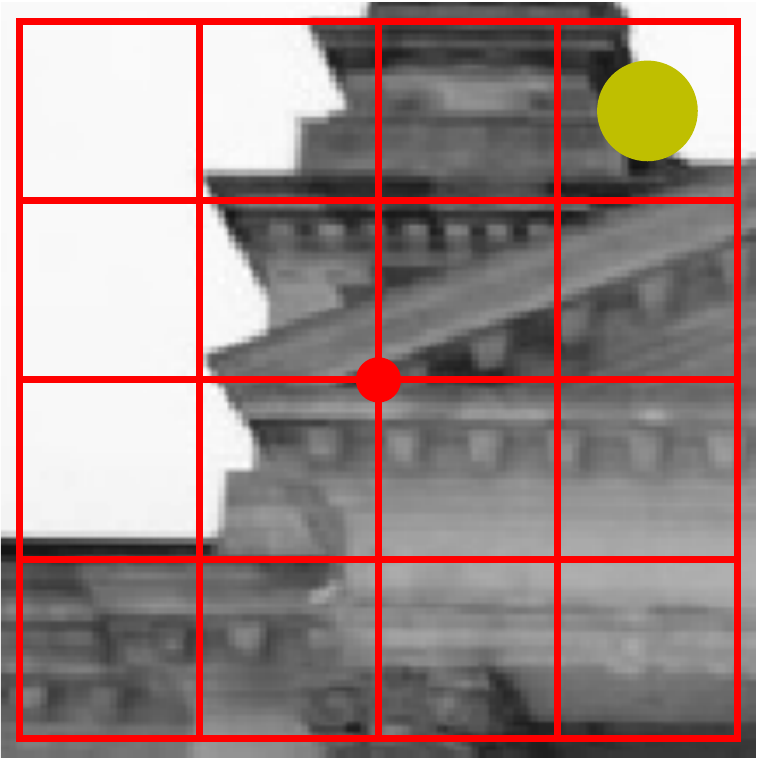} &
\includegraphics[width=\imsize]{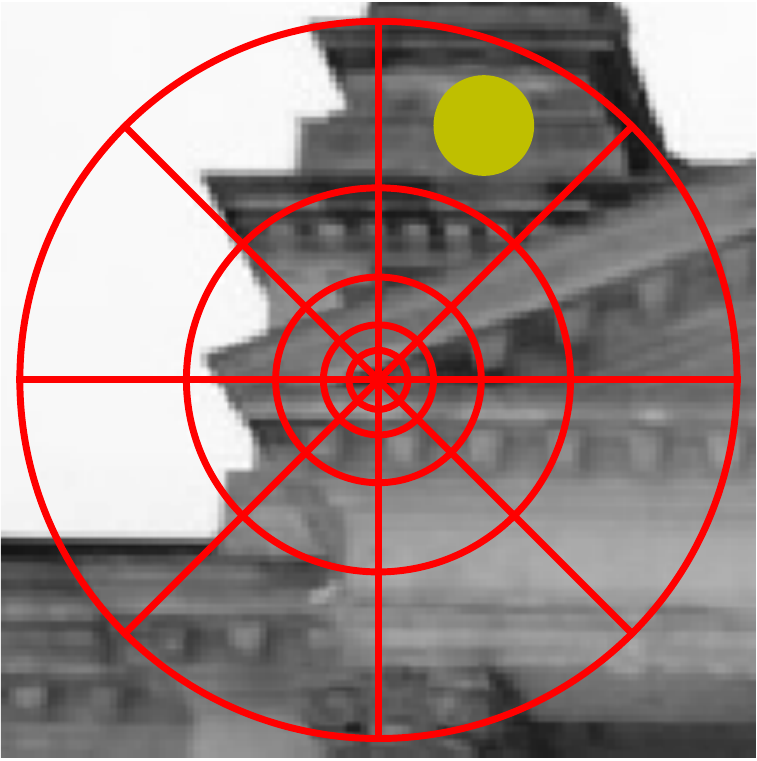} &
\includegraphics[width=\imsize]{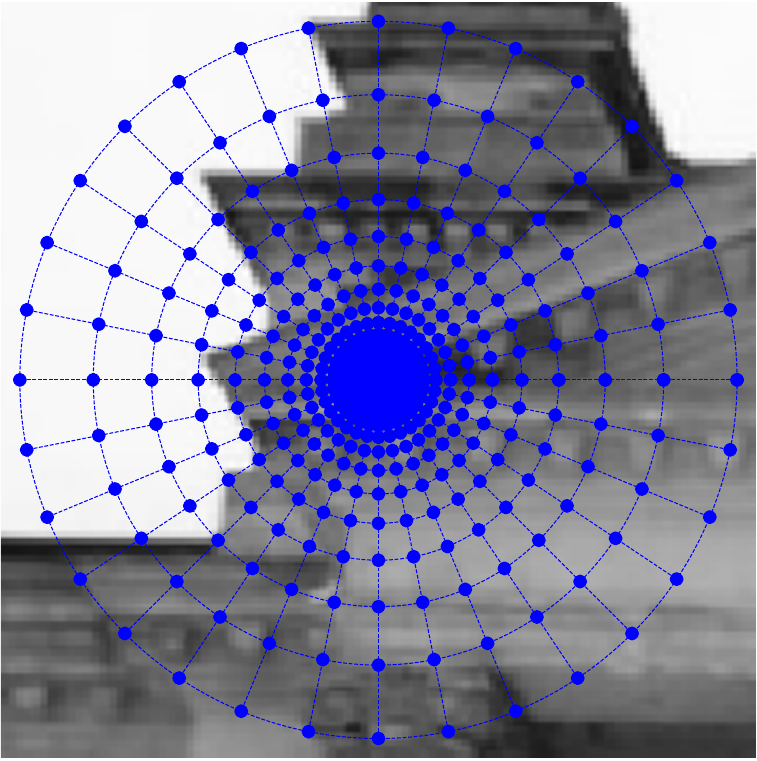} &
\includegraphics[width=\imsize]{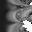} \\
  &
{\footnotesize (a) Cartesian} &
{\footnotesize (b) Log-Polar} &
{\footnotesize (c) Log-Polar} &
{\footnotesize (d) Log-Polar} \\
&
{\footnotesize Pooling (SIFT)}\vspace{0.5mm} &
{\footnotesize Pooling} &
{\footnotesize Sampling} &
{\footnotesize Patch} \\
\end{tabular}
\vspace{-1mm}
\caption{{\bf Pooling vs Sampling.} (a,b) The red patterns depict the regions that most descriptors use to {\it pool} features computed on a cartesian pixel-grid. The size of the pattern depends on the local scale, and we show three versions. Under large scale changes, many regions of the cartesian and log-polar grids, such as the ones highlighted by the yellow dots, can no longer be put in correspondence. (c) By contrast, we first resample the patch according to the patterns shown in blue (32 $\times$ 32). Their size also depends on the scale. (d) Even though the scales vary from 1 to 4, the resulting log-polar patches are all fairly similar, notably near the center of the feature location, depicted by the left side of the patch.}
\label{fig:sampling}
\vspace{-3mm}
\end{figure}

Polar and log-polar representations have been extensively used in computer vision to aggregate local information, because they are robust to small changes in scale and rotation. Traditional hand-crafted patch descriptors typically consist of two stages: feature {\em extraction} and feature {\em pooling}. First, image measurements such as gradients are computed for every pixel. Then, they are aggregated over small regions around the point given its location, orientation, and scale.
SIFT, for instance, aggregates features (histograms of gradient orientations) over 4$\times$4 cells around the keypoint; see \fig{sampling}.

Several descriptors aggregate features over polar or log-polar regions. GLOH~\cite{Miko04b} computes SIFT over a log-polar grid and then reduces the dimensionality by PCA. Daisy~\cite{Tola08} aggregates oriented image gradients over a polar grid using a Gaussian kernel with a size proportional to the distance between the keypoint and the grid point, to bypass aliasing effects. The seminal Shape Contexts paper~\cite{Belongie02} introduces a descriptor for object recognition by picking points on the contour of a shape and histogramming the location of each point relative to every other point over log-polar bins. Local Self-Similarities (LSS)~\cite{Shechtman07} proposes a technique to match different image modalities by measuring internal self-similarities over the regions determined by a log-polar grid. Winder and Brown~\cite{Winder07} study many pooling configurations within a framework similar to Daisy and find log-polar to be optimal among their choices. Several binary descriptors, such as BRISK~\cite{Leutenegger11} or FREAK~\cite{Alahi12}, rely on sampling patterns over similarly-defined grids to compute intensity differences and extract the features.

Note that all of these methods define polar or log-polar regions for feature {\em pooling}, that is, the pixel-wise features are always computed in cartesian space, and it is only their aggregation that takes place in log-polar space. As shown in \fig{sampling}, this is drastically different to our approach, which consists in warping the raw pixel data and use that representation to learn scale-invariant models.

\section{Method}
\label{sec:method}

\setlength{\fboxsep}{0pt} %
\newcommand{\imsizebigone}{4.2cm}%
\newcommand{\imsizebigtwo}{4.2cm}%
\newcommand{\imsizesmall}{0.96cm}
\begin{figure*}
\renewcommand{\arraystretch}{0.8}
\renewcommand{\tabcolsep}{0.7mm}
\centering
\begin{tabular}{@{}cccc@{}}
\includegraphics[height=\imsizebigone]{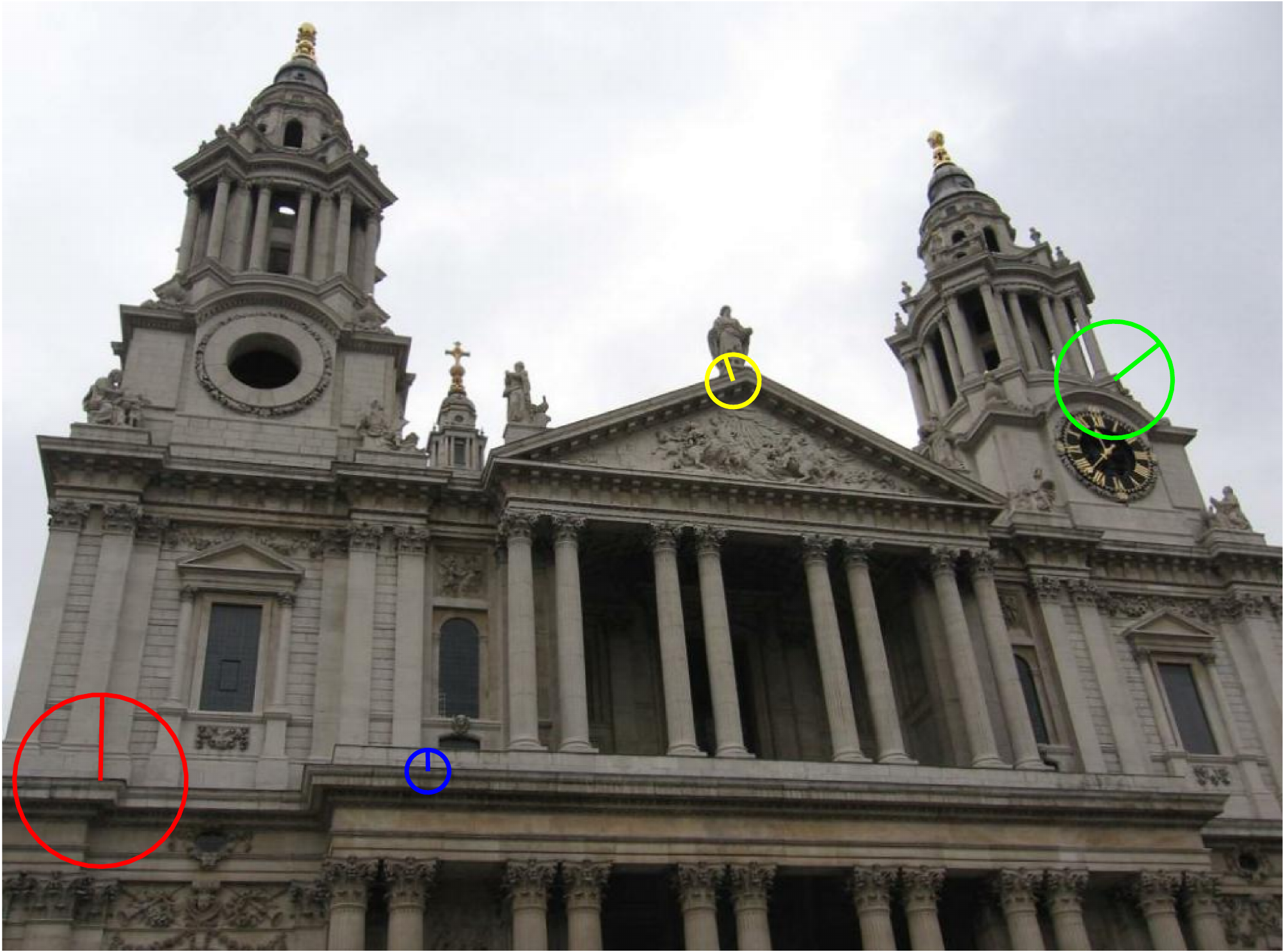} & 
\begin{tabular}[b]{@{}cc@{}}%
\fcolorbox{blue}{white}{\includegraphics[height=\imsizesmall]{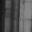}} &
\fcolorbox{blue}{white}{\includegraphics[height=\imsizesmall]{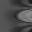}} \\
\fcolorbox{red}{white}{\includegraphics[height=\imsizesmall]{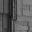}} &
\fcolorbox{red}{white}{\includegraphics[height=\imsizesmall]{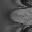}} \\
\fcolorbox{green}{white}{\includegraphics[height=\imsizesmall]{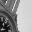}} &
\fcolorbox{green}{white}{\includegraphics[height=\imsizesmall]{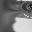}} \\
\fcolorbox{yellow}{white}{\includegraphics[height=\imsizesmall]{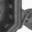}} &
\fcolorbox{yellow}{white}{\includegraphics[height=\imsizesmall]{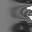}} \\
\end{tabular} &
\includegraphics[height=\imsizebigtwo]{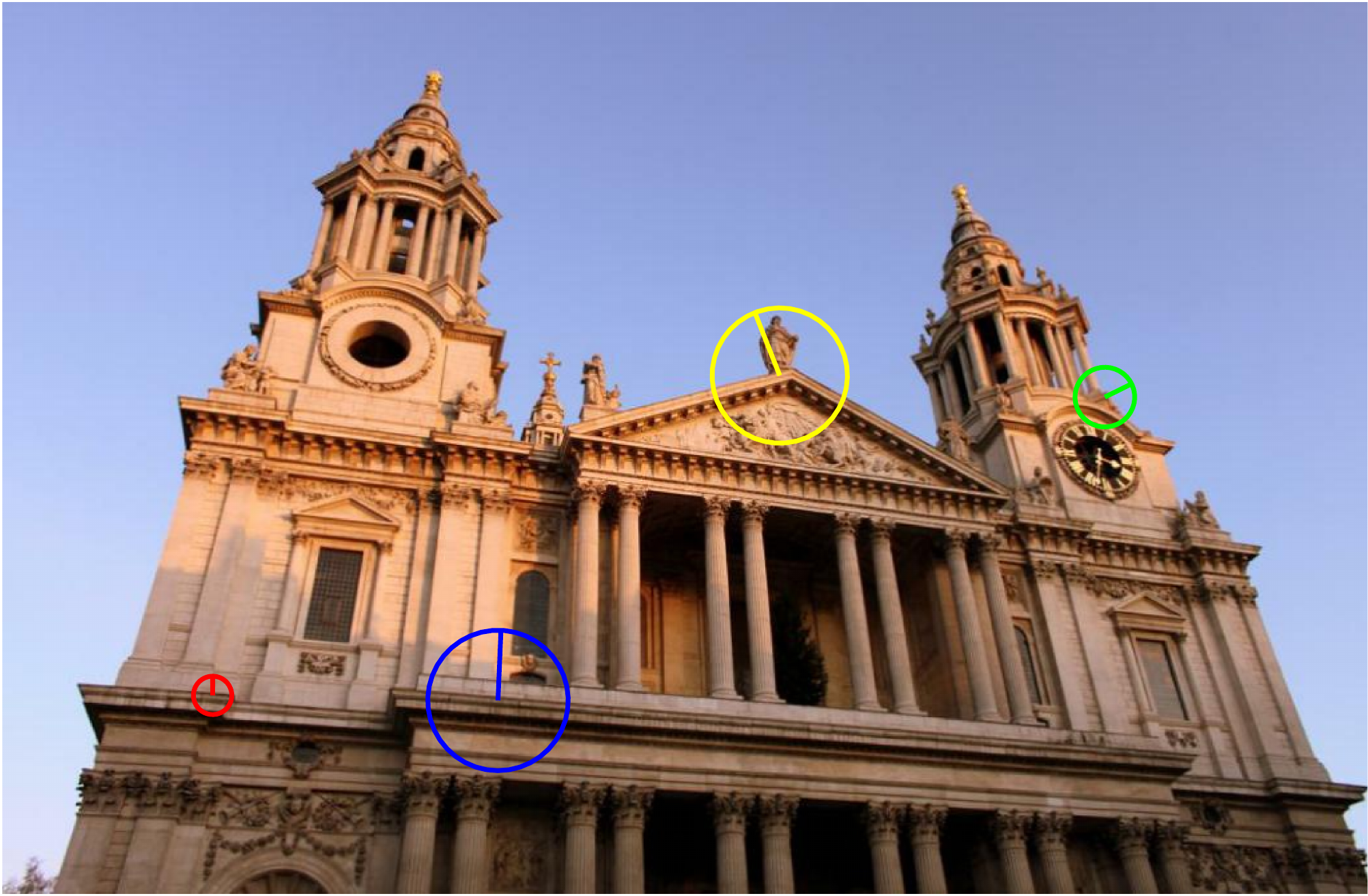} & 
\begin{tabular}[b]{@{}cc@{}}%
\fcolorbox{blue}{white}{\includegraphics[height=\imsizesmall]{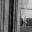}} &
\fcolorbox{blue}{white}{\includegraphics[height=\imsizesmall]{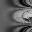}} \\
\fcolorbox{red}{white}{\includegraphics[height=\imsizesmall]{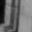}} &
\fcolorbox{red}{white}{\includegraphics[height=\imsizesmall]{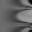}} \\
\fcolorbox{green}{white}{\includegraphics[height=\imsizesmall]{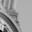}} &
\fcolorbox{green}{white}{\includegraphics[height=\imsizesmall]{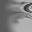}} \\
\fcolorbox{yellow}{white}{\includegraphics[height=\imsizesmall]{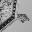}} &
\fcolorbox{yellow}{white}{\includegraphics[height=\imsizesmall]{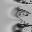}} \\
\end{tabular} \\
{\small (a)} &
{\small (b)} &
{\small (c)} &
{\small (d)} \\
\end{tabular}
\vspace{-3mm}
\caption{{\bf Cartesian vs Log-Polar.} (a,c) Two images taken from different viewpoints with four pairs of corresponding keypoints, denoted by their color. (b,d) Patches around these keypoints extracted with their estimated scale and orientation, with $\lambda$=16, similarly color-coded. On each column, we show cartesian patches on the left and log-polar patches on the right. While cartesian patches can look very different, log-polar ones remain similar. This is particularly visible for the red keypoint, whose scale estimates are very different in the two images.}
\vspace{-3mm}
\label{fig:logpol_examples}
\end{figure*}

First, we describe our sampling scheme in \refsec{sampling} and, then, our network architecture and training strategy in \refsec{network}. For the purposes of this section, we assume that the training data consists of pairs of keypoints across two images that are in correspondence in terms of location and orientation, but not necessarily scale. %
The actual procedure used to generate the training data is described in \refsec{dataset}.

\subsection{Log-Polar Sampling}
\label{sec:sampling}

\setlength{\abovedisplayskip}{6pt}
\setlength{\belowdisplayskip}{6pt}
As in most papers about learning descriptors~\cite{Han15,Zagoruyko15,Simo-Serra15,Balntas16b,Tian17,Mishchuk17}, we use SIFT keypoints~\cite{Lowe04}. Given an image $\bI$ of size $H \times W$, a keypoint $\kp_i$ on $\bI$ is fully described by its center coordinates $\left( x_i, y_i \right)$, its scale $\scale_i\in\mathbb{R}_+$, and its orientation $\ori_i \in [0, 2\pi)$. We use a Polar Transformer Network (PTN)~\cite{Esteves18} to extract a $L \times L$ patch around keypoint $\kp_i$. To this end, we rely on the following coordinate transform: %
\begin{align}
x_i^s & = x_i + e^{log(r_i) x_i^t/W} cos\left(\varphi_i\right) \; , \\
y_i^s & = y_i + e^{log(r_i) x_i^t/W} sin\left(\varphi_i\right) \; . \nonumber
\label{eq:polar}
\end{align}
Variables $(x_i^s, y_i^s)$ denote source coordinates and $(x_i^t, y_i^t)$ target coordinates, after the transform. The coordinate origin is centered on $\left( x_i, y_i \right)$, the angle is
$\varphi_i=\frac{\ori_i+2\pi y_i^t}{H}$,
and the radius $r_i$ is given by $\frac{\mult}{2} \scale_i$, where $\mult$ is a factor that converts the SIFT scale to image pixels\footnote{Given the convention followed by OpenCV, $\mult=12$ denotes the scale multiplier of SIFT. One can extract larger image regions setting $\mult > 12$.}.
Finally, we construct the warped patches by looking up the intensity values in image $\bI$ at coordinates
$\left( x_i^t, y_i^t \right)$
with bilinear interpolation, as done in~\cite{Esteves18}. This process is illustrated in \fig{sampling}.
We denote patches extracted in this way as {\bf LogPol}. For comparison purposes, we also consider the standard cartesian approach, using Spatial Transformer Networks (STN)~\cite{Jaderberg15} on a regularly spaced sampling grid, defined by
\begin{align}
x_i^t & = x_i + x_i^s cos\left(\theta_i\right) \sigma_i / W - y_i^s sin\left(\theta_i\right) \sigma_i / H\; , \\
y_i^t & = y_i + x_i^s sin\left(\theta_i\right) \sigma_i / W + y_i^s sin\left(\theta_i\right) \sigma_i / H\; .\nonumber
\label{eq:cartesian}
\end{align}
We denote these patches as {\bf Cart}. 
Note that STN and PTN were designed to facilitate whole-image classification by allowing deep networks to manipulate data spatially, thus removing the burden of learning spatial invariance from the classifier.
This is not applicable here: we only use their respective samplers, which
allow us to efficiently sample the images with in-line data augmentation at a negligible computational cost by applying small perturbations
when extracting the patches.

\begin{figure*}[!t]
  \centering
  \begin{tabular}{c}
     \includegraphics[height=2cm]{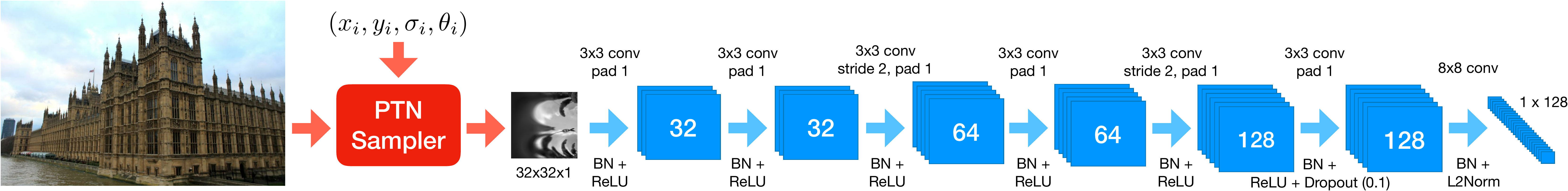} \\
  \end{tabular}
\vspace{-3mm}
\caption{{\bf Network architecture.} We extract patches size 32$\times$32 in-line with a sampler (pictured: PTN) on the desired keypoints. This enables data augmentation techniques. The patches are then given to a network which produces descriptors size 128.}
\vspace{-3mm}
\label{fig:hardnet}
\end{figure*}

The following properties of log-polar patches distinguish them from cartesian ones: 
\begin{itemize}[noitemsep,nolistsep]
  \item Rotations in cartesian space correspond to shifts on the polar axis in log-polar space (rotation equivariance).
  \item Peripheral %
    regions are undersampled, which means that paired patches look similar to the eye even under drastic scale changes (scale equivariance).
\end{itemize}
This phenomena is illustrated in \fig{logpol_examples}. Note how the log-polar representation facilitates visual matching even when scales are mismatched. Our approach is predicated on leveraging this information effectively with the deep networks and training framework introduced in the next section.

\subsection{Network Architecture and Training}
\label{sec:network}

To extract patch descriptors, we use a HardNet~\cite{Mishchuk17} architecture.
As shown in~\fig{hardnet}, our
network has seven convolutional layers and takes as input grayscale patches of size 32$\times$32.
Input patches are pre-processed with Instance Normalization ~\cite{Ulyanov17}.
Feature maps are zero-padded after each convolutional layer, and we use strided convolutions instead of pooling layers.  Each convolution is followed by a ReLU and Batch Normalization, but the last convolution layer omits the ReLU. We apply dropout regularization with a rate of 0.1 after the last ReLU. The final convolutional layer is followed by Batch Normalization and $l_2$ normalization. The output of the network is a descriptor of unit length and size 128.
We found this to be a good compromise between descriptor size and performance.

The standard way to train such networks is in a siamese configuration, with two copies of the network, sharing weights. Among the many loss formulations that have have been proposed~\cite{Simo-Serra15,Balntas16b,Keller18,He18b}, we use the triplet loss of~\cite{Balntas16b}, as in~\cite{Mishchuk17}.
To build the required triplets, we consider a collection of patch pairs $\{\pa_k,\pb_k\}$ which contain two different views of a 3D point, where $k = 1 \dots K$, with $K$ denoting the batch size. We systematically check that the 3D points in a given batch are unique, so that $\pa_i$ and $\pb_j$ only correspond if $i = j$.
We denote their respective descriptors as $\{\da_k,\db_k\}$.
We then mine negative samples with the `hardest-in-batch' procedure of~\cite{Mishchuk17}. Specifically, we build a pairwise distance matrix $\mathbf{D}_{i,j} = d(\da_i, \db_j)$, $i,j \in [1,K]$, where $d(\da_i,\db_j)$ is the Euclidean distance between descriptors $\da_i$ and $\db_j$ if $i \neq j$, and an arbitrarily large value otherwise. We denote the hardest negative sample for $\pa_k$, \ie, the one with the smallest distance, as $\pb_{k_{min}}$, and the hardest negative sample for $\pb_k$ as $\pa_{k_{min}}$. We consider both $\pa_k$ and $\pb_k$ as possible anchors, for all $k$. Denoting a triplet with anchor ($A$), positive ($+$) and negative ($-$) patches as $(A,+,-)$, we form triplet $k$ taking the hardest negative example, \ie $\{\pa_k,\pb_k,\pb_{k_{min}}\}$ if $d(\pa_k,\pb_{k_{min}}) < d(\pb_k,\pa_{k_{min}})$ and $\{\pb_k,\pa_k,\pa_{k_{min}}\}$ otherwise.
We then take the loss to be
\begin{equation*}
  \loss(\danc, \dpos, \dneg) = \sum_{k=1}^K
    \max \left(
      0,
      1  + | \danc_k - \dpos_k |^2 - | \danc_k - \dneg_k |^2
    \right).
\label{eq:loss_desc}
\end{equation*}
We set the batch size $K$ to 1000.
For optimization we use Stochastic Gradient Descent (SGD) with a learning rate of 10%
, momentum of 0.9, weight decay $10^{-4}$, and decay the learning rate linearly to zero within a set number of training epochs \cite{Mishchuk17}.
Sampling the patches in-line allows us to apply data augmentation at training time, jittering the orientation of each anchor keypoint by $\Delta\ori \sim \mathcal{N}(0,25)$ degrees.
Our implementation uses Pytorch as a back-end. Code, models and training data are all available.\footnote{{\scriptsize \url{https://github.com/cvlab-epfl/log-polar-descriptors}}}

\section{Experiments}
\label{sec:experiments}

In \refsec{dataset}, we introduce the dataset we built to train scale-invariant descriptors, because there is currently none available for this purpose. We then compare ourselves to the state of the art on it.
In Sections \ref{sec:exp_hpatches}, \ref{sec:exp_amos} and \ref{sec:exp_challenge}, we benchmark our models on three publicly available datasets: HPatches~\cite{Balntas17}, AMOS patches~\cite{Pultar19}, and the CVPR'19 Photo Tourism image matching challenge~\cite{imwchallenge2019}.
As baselines, we consider: SIFT~\cite{Lowe04}, 
TFeat~\cite{Balntas16b},
L2-Net~\cite{Tian17}, HardNet~\cite{Mishchuk17}, and GeoDesc~\cite{Luo18}.\footnote{We use OpenCV for SIFT, and public implementations for the rest.}
For our own method we consider descriptors learned with either cartesian or log-polar patches.

\subsection{Results on the New Dataset}
\label{sec:dataset}

\renewcommand{\arraystretch}{1}
\renewcommand{\tabcolsep}{3mm}

\begin{table*}
	\small
	\begin{center}
		\begin{tabular}{@{}lcccccccccc@{}}
			\toprule
			\multirow{2}{*}{Sequence} & \multirow{2}{*}{SIFT} & \multirow{2}{*}{TFeat} & \multirow{2}{*}{L2-Net} & \multirow{2}{*}{Geodesc} & \multirow{2}{*}{HardNet} & \multicolumn{2}{c}{Ours ($\lambda=12$)} & Ours ($\lambda=96$) \\
			& & & & & & Cart & LogPol & LogPol\\
			\midrule
      		`british\_museum'           & 5.91 & 3.53 & 3.52 & 4.30 & 3.21 & 2.17 & 2.18 & {\bf 0.96} \\
			`florence\_cathedral\_side' & 4.36 & 1.30 & 0.51 & 2.13 & 0.40 & 0.23 & 0.23 & {\bf 0.20} \\
			`lincoln\_memorial\_statue' & 2.89 & 4.32 & 2.28 & 2.61 & 1.65 & 1.30 & 1.31 & {\bf 0.91} \\
			`milan\_cathedral'          & 7.08 & 1.98 & 1.48 & 1.86 & 0.35 & 0.19 & 0.12 & {\bf 0.07} \\
			`mount\_rushmore'           & 18.71 & 11.94 & 2.52 & 2.27 & 0.43 & 0.42 & 0.32 & {\bf 0.22} \\
			`reichstag'                 & 2.22 & 0.44 & 0.30 & 0.42 & 0.21 & 0.19 & 0.19 & {\bf 0.09} \\
			`sagrada\_familia'          & 9.01 & 2.41 & 0.85 & 1.08 & 0.27 & 0.21 & 0.19 & {\bf 0.03} \\
			`st\_pauls\_cathedral'      & 8.64 & 2.01 & 1.48 & 2.45 & 0.68 & 0.42 & 0.46 & {\bf 0.20} \\
			`united\_states\_capitol'   & 8.67 & 3.90 & 2.64 & 5.43 & 1.60 & 1.33 & 0.98 & {\bf 0.53} \\
			\midrule
			Average                     & 7.50 & 3.54 & 1.73 & 2.51 & 0.98 & 0.72 & 0.67 & {\bf 0.36} \\
			\bottomrule
		\end{tabular}
	\end{center}
	\vspace{-3mm}
  \caption{{\bf FPR95 on our new dataset.} We benchmark our models against the baselines with patches extracted at the SIFT scale, $\lambda=12$.
    We also show that log-polar models are able to leverage much larger support regions, using $\lambda=96$. By contrast, with cartesian patches performance degrades as we increase the support region, as we demonstrate in the ablation study of \tbl{results_lambda}.
	}
	\vspace{-3mm}
	\label{tbl:results_patch_all}
\end{table*}

Nearly all learned descriptors rely on the dataset of~\cite{Brown10} for training, which provides patches extracted over different viewpoints for three different scenes. Correspondences were established from SfM reconstructions and SIFT. They are thus biased towards keypoints that can be matched with SIFT, \ie, commensurate in terms of scale. In order to learn scale-invariant descriptors under real-world conditions, we require patches extracted at non-corresponding scales, for which we need the original images, which are not provided by~\cite{Brown10}. Other datasets, such as~\cite{Mikolajczyk05} or~\cite{Balntas17}, provide images along with homographies for correspondence,
but focus on affine transformations and are much too small to train deep networks effectively. Therefore we collected a new dataset for training purposes. In the remainder of this section, we detail how we created it and then report our results on it.

\vspace{-4mm}
\subsubsection{Creating the Dataset}
\label{sec:dataset_data}

We applied COLMAP~\cite{Schoenberger16a}, a state-of-the-art SfM framework, over large collections of photo-tourism images originally collected by~\cite{Heinly15}.
These images show drastic changes in terms of viewpoint, illumination, and other imaging properties, which is crucial to learn invariance~\cite{Yi16b}.
In addition to sparse reconstructions, COLMAP provides
dense correspondences in the form of depth maps.
We used them to generate training data by randomly selecting a pair of images $\bI_i$ and $\bI_j$, extracting SIFT keypoints for both, and using the depth maps to build ground truth correspondences.
To do this we projected each keypoint from one image to the other using the estimated poses and depth maps. We took a correspondence $(m,n)$ to be valid if the projection of keypoint $m$ in image $i$ falls within 1.5 pixels of keypoint $n$ in image $j$. We performed a bijective check to ensure one-to-one correspondences. We applied this projection in a cycle, from $i$ to $j$ and back to $i$, to ensure that the depth estimates are consistent across both views, and discarded the putative correspondence otherwise.
Points which fall in occluded areas were likewise discarded. Note that we only check for corresponding {\em locations}, but not scales: in this manner we are collecting SIFT keypoints with non-matching scales whose distribution comes from real-world data.
We also require the orientations to be compatible across views.
To guarantee this we use the ground truth camera poses to compute the difference between orientation estimates and filter out keypoint matches outside 25$^o$, as in~\cite{Brown10}.
Finally, we suppress pairs of keypoints closer than 7 pixels to each other, to exclude patches with large overlaps,
which would be particularly problematic for cartesian patches.%

We can similarly use the ground truth to warp the scale across images, which we do in order to estimate the frequency of inaccurate scale estimates. Given a correspondence $(m,n)$ comprised of two keypoints with scales $(s_i^m,s_j^n)$, we warp the scale from image $i$ to image $j$ to obtain $\hat{s}_i^m$, and compute the scale difference ratio as $r = \frac{max(\hat{s}_i^m, s_j^n)}{min(\hat{s}_i^m, s_j^n)}$, so that $r \geq 1 $, with 1 encoding perfect scale correspondence. We histogram this ratio and %
use it to evaluate each method under scale changes, as depicted in
\fig{fpr95bins}.

\renewcommand{\arraystretch}{0}
\renewcommand{\tabcolsep}{0.3mm}
\renewcommand{\imsize}{3.6cm}
\begin{figure}
\centering
\begin{tabular}{cccc}
& \hspace{6mm}{\scriptsize Scale change} & \hspace{0cm}{\scriptsize Scale change} & \vspace{1mm} \\
\raisebox{3em}{\rotatebox{90}{\raisebox{0.3em}{{\scriptsize Orientation change}}}} &
\includegraphics[height=\imsize]{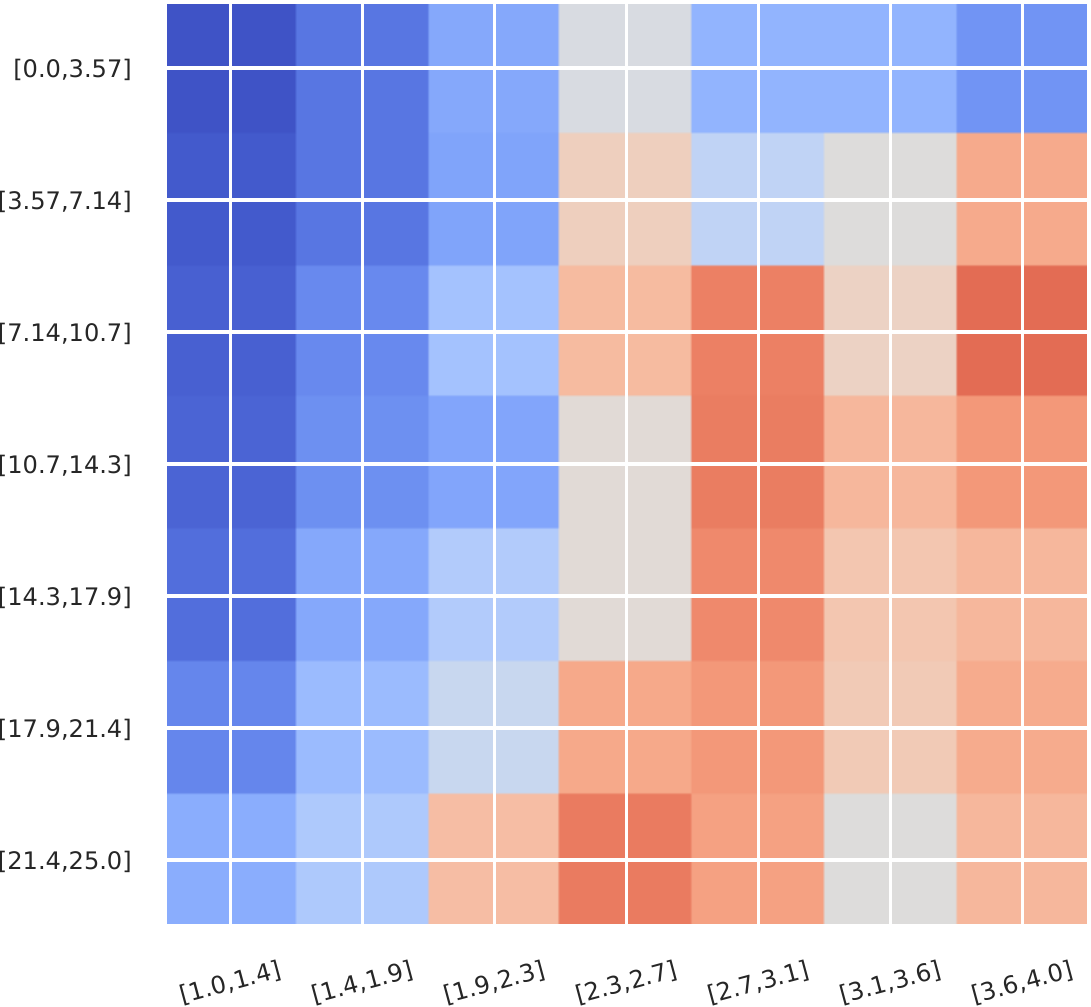} &
\includegraphics[height=\imsize]{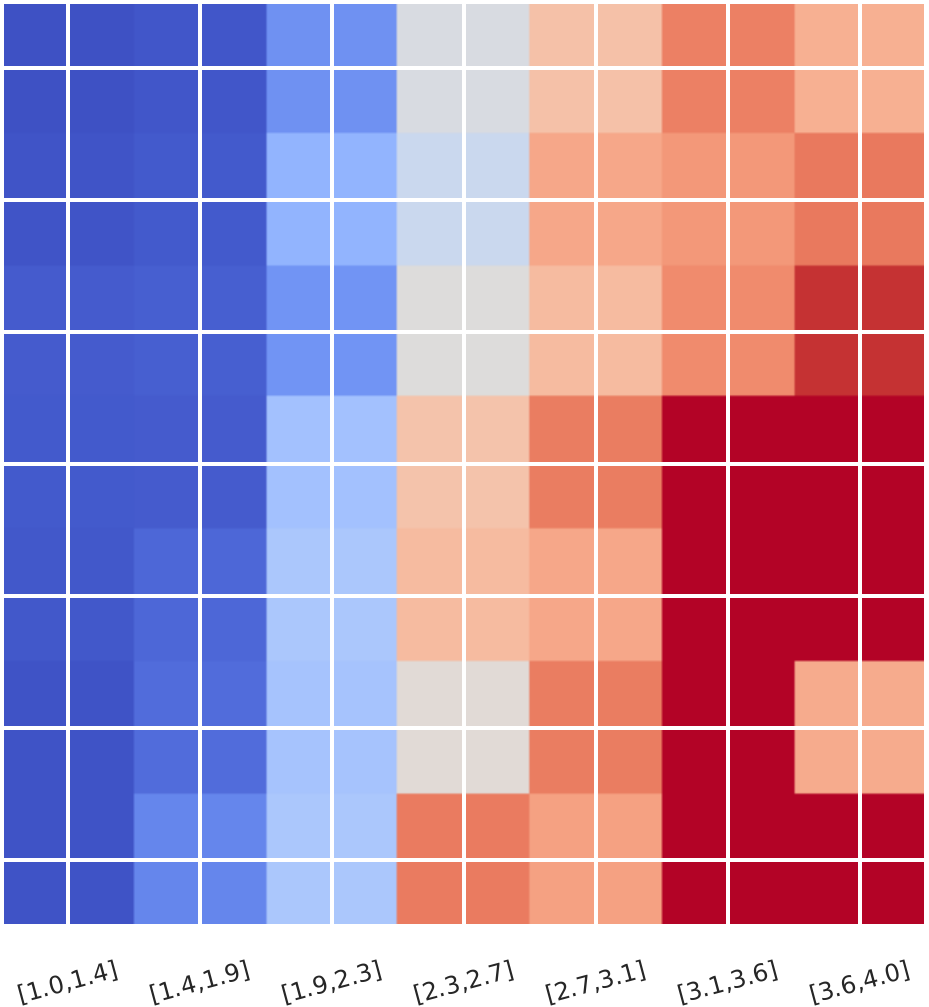} & \vspace{2mm}
\includegraphics[height=\imsize]{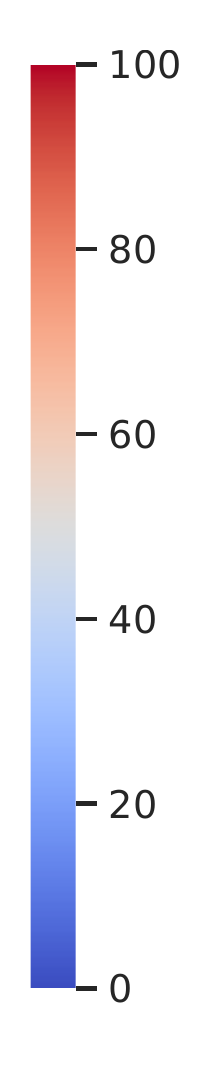} \\
& \hspace{6mm}{\footnotesize (a) SIFT} & {\footnotesize (b) L2-Net}\vspace{2mm} \\
\raisebox{3em}{\rotatebox{90}{\raisebox{0.3em}{{\scriptsize Orientation change}}}} &
\includegraphics[height=\imsize]{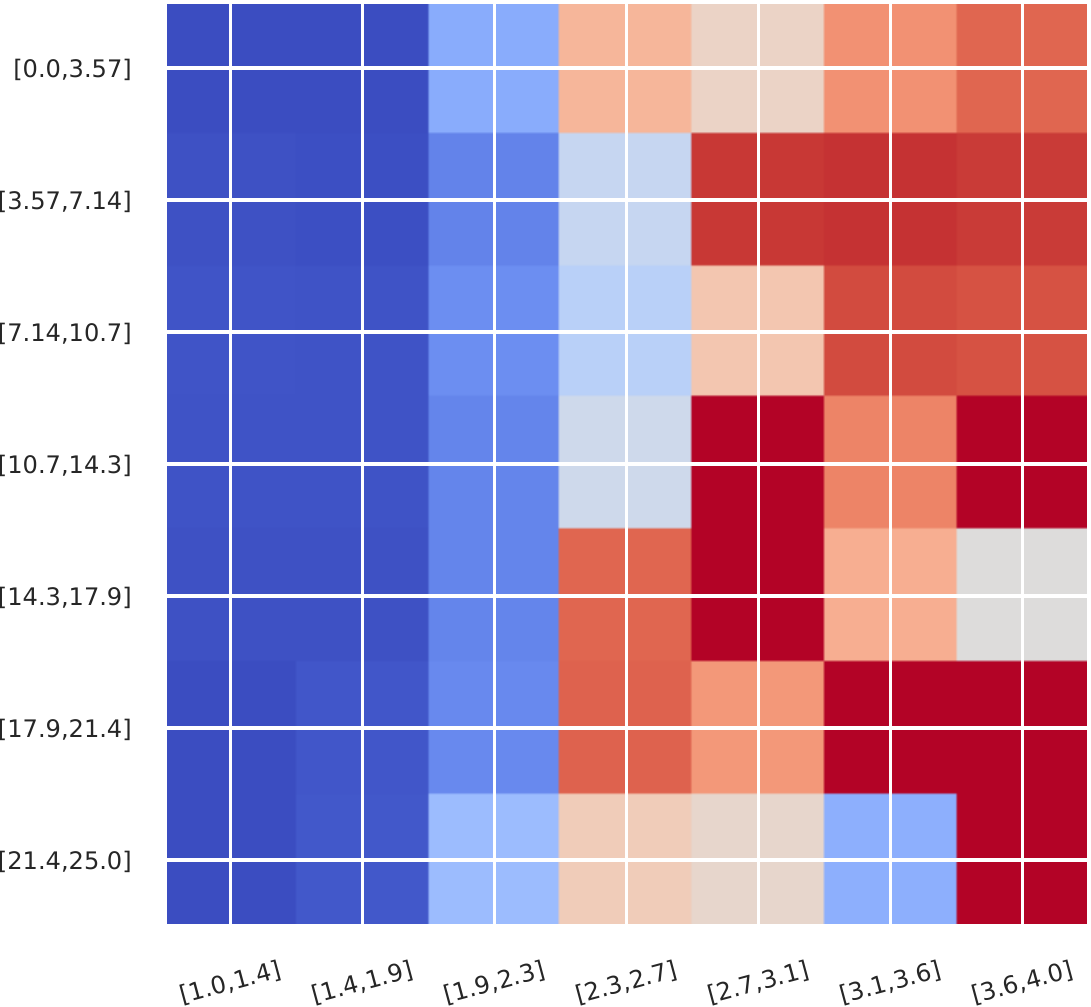} &
\includegraphics[height=\imsize]{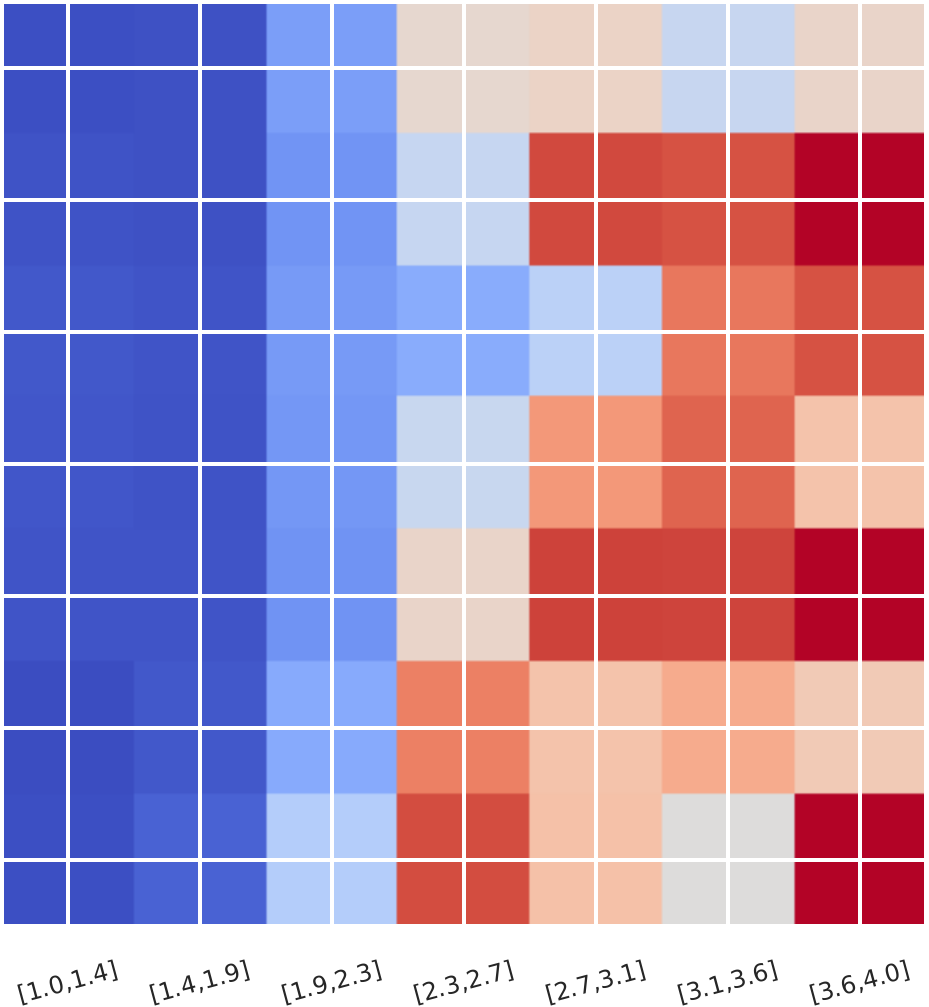} & \vspace{2mm}
\includegraphics[height=\imsize]{figs/fpr95_per_bin_linear/colorbar-big.pdf} \\
& \hspace{6mm}{\footnotesize (c) Ours-Cart ($\lambda=12$)} & {\footnotesize (d) HardNet}\vspace{2mm} \\
\raisebox{3em}{\rotatebox{90}{\raisebox{0.3em}{{\scriptsize Orientation change}}}} &
\includegraphics[height=\imsize]{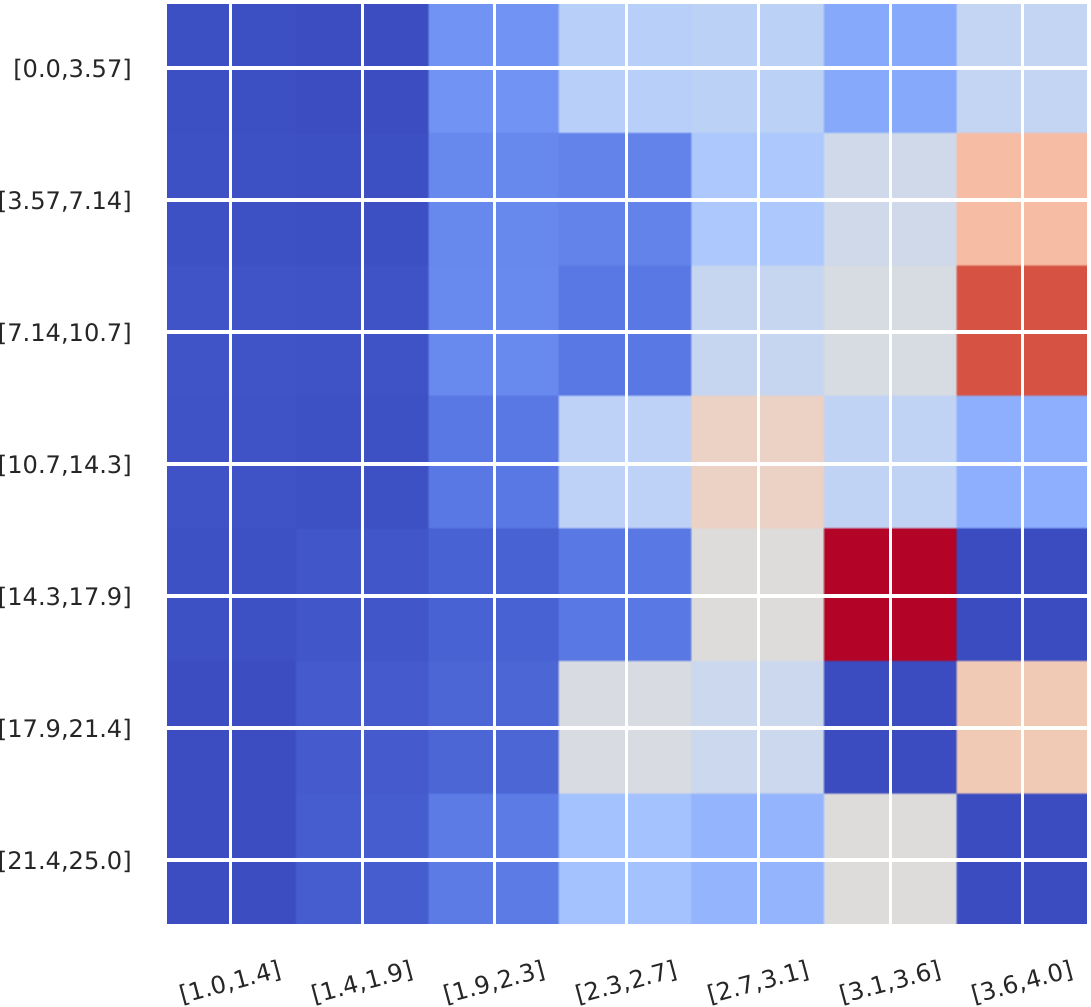} &
\includegraphics[height=\imsize]{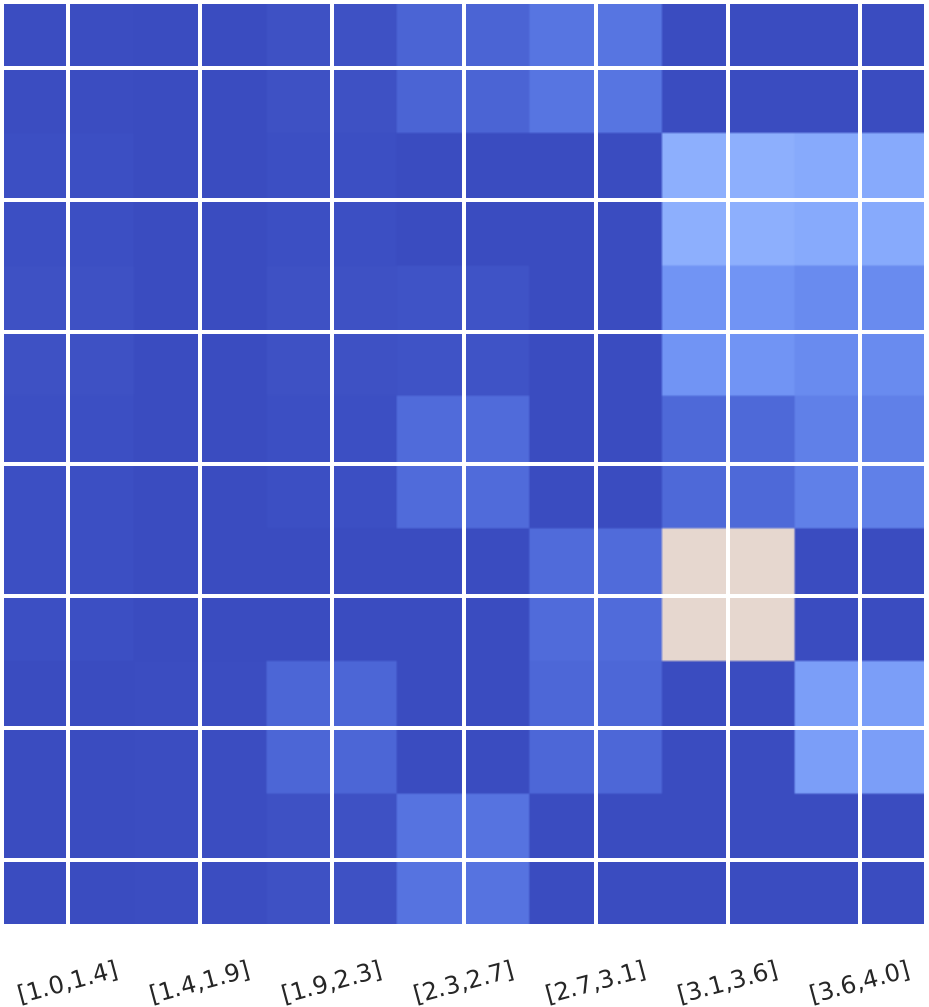}  & \vspace{2mm}
\includegraphics[height=\imsize]{figs/fpr95_per_bin_linear/colorbar-big.pdf} \\
& \hspace{6mm}{\footnotesize (e) Ours-LogPol ($\lambda=12$)} & {\footnotesize (f) Ours-LogPol ($\lambda=96$)}\vspace{2mm} \\
\end{tabular}

\caption{{\bf FPR95 vs Scale and orientation changes.} We break down the results of \tbl{results_patch_all}, histogramming them by the error in the keypoint detection stage. Orientation misdetections increase top to bottom, up to 25$^o$. Scale misdetections increase left to right, up to 4x.
(a,b,d) All baselines degrade quickly under scale changes.
(c) Training deep networks with cartesian patches with scale changes is not sufficient.
(e,f) By contrast, our log-polar representation enables them to learn scale invariance.
Note that some bins are sparsely populated, which explains sudden discontinuities.}
\vspace{-1mm}
\label{fig:fpr95bins}
\end{figure}

We select 11 sequences for training, and 9 for testing. Please refer to the supplementary material for details. We split the training sequences into training and validation sets in a per-image basis, with a 70:30 ratio. Images are downsampled to a maximum height or width of 1024 pixels, which is the resolution that we extract keypoints at, and mirror-padded to 1500$\times$1500 to alleviate boundary effects. To obtain patches in cartesian space, we sample the image at the desired keypoints with STN. For log-polar patches we use PTN over a support region commensurate with STN; see \secref{sampling}. We also consider larger patches, increasing $\lambda$.
We generate up to 1000 correspondences for each image pair, and extract the patches from the images on the fly.

Training requires negative samples, that is, points {\em not} in correspondence. Finding negatives is easy when a SfM reconstruction is available, as done in~\cite{Brown10}, ensuring that keypoints are stable across all views. This not feasible in our case.
Instead, we generate training samples from a single image pair at a time. Specifically, we take one image pair from each of the 11 sequences and use it to fill roughly $1/11$th of each training batch. We can then perform negative mining over the entire batch, as outlined in \refsec{network}.

\vspace{-2mm}
\subsubsection{Patch Correspondence Verification}
\label{sec:exp_patches}

In this section we evaluate performance in terms of patch matching over the test sequences. We extract descriptors for SIFT keypoints with corresponding locations, but using their original scales, which are not always in correspondence.
We train our networks with cartesian and log-polar patches, keeping all other settings identical. We use the standard metric in patch matching benchmarks, FPR95, \ie, the False Positive Rate at 95\% True Positive recall.
For the baselines, we extract patches at the SIFT scale, \ie, $\lambda=12$. We also consider $\lambda > 12$ for log-polar patches.
We report the results in \tbl{results_patch_all} and discuss them below.

\vspace{-4mm}
\paragraph{Comparison to the state of the art.}
Our models trained with log-polar patches deliver the best performance on each sequence, followed by our models trained on cartesian patches, and then HardNet. Remarkably, we achieve our best results with $\lambda=96$, which corresponds to patches much larger than those best-suited for traditional descriptors, extracted with $\lambda=12$, a fact that we will examine more closely below. Note the small gap between HardNet and Ours-Cartesian, which is due to the innate differences between datasets and training the latter with mismatched scales. The other baselines perform significantly worse. 
\vspace{-4mm}
\paragraph{Performance under large scale mismatches.} In~\fig{fpr95bins} we break down the results of \tbl{results_patch_all} in terms of orientation and scale mismatches.
Note how models trained on log-polar representations can tolerate a wide range of scale mismatches. Our results show a negligible drop in performance under scale changes up to 2-3x, and remain useful even at 3-4x. All baselines degrade significantly under scale changes of 2x and become essentially useless beyond that.
Note that this invariance is made possible by leveraging log-polar representations and cannot be achieved by simply exposing the models to cartesian patches exhibiting scale changes, 
as evidenced by the performance of Ours-Cartesian shown in \fig{fpr95bins}-\et{(c).}
Finally, remember that this data has been collected from real-world settings with unreliable scale detection.
In other words, our models allow us to retrieve {\em more correspondences} without changing the detector.

\vspace{-4mm}
\paragraph{Increasing the size of the support region.} As shown in \fig{logpol_examples}, patches extracted with log-polar sampling are remarkably similar across different scales, because scale changes correspond to shifts in the horizontal dimension. This representation is not only easier to interpret visually, but also easier to learn invariant models with. Moreover, oversampling the immediate neighbourhood of the point allows us to leverage larger support regions, because the effect of occlusions and background motion in log-polar patches is smaller than in their cartesian counterparts.
We demonstrate this by training models for different values of $\lambda$, and report the results in \tbl{results_lambda}.
Our models are able to exploit support regions much larger than cartesian-based approaches.
We see performance flatten out at $\lambda=96$, and observe boundary issues beyond that point, so we use this value for all experiments in the paper.
Note how the radius of the circle determining the support region is {\em 8 times larger} than the optimal value for cartesian patches, and its area {\em 64 times larger}. Note that we use an identical architecture, which can only leverage this information effectively thanks to the log-polar representation.

\vspace{-2mm}
\subsubsection{Image-level Patch Retrieval}
\label{sec:exp_ranking}

\begin{figure}
\centering

\includegraphics[width=\linewidth, clip]{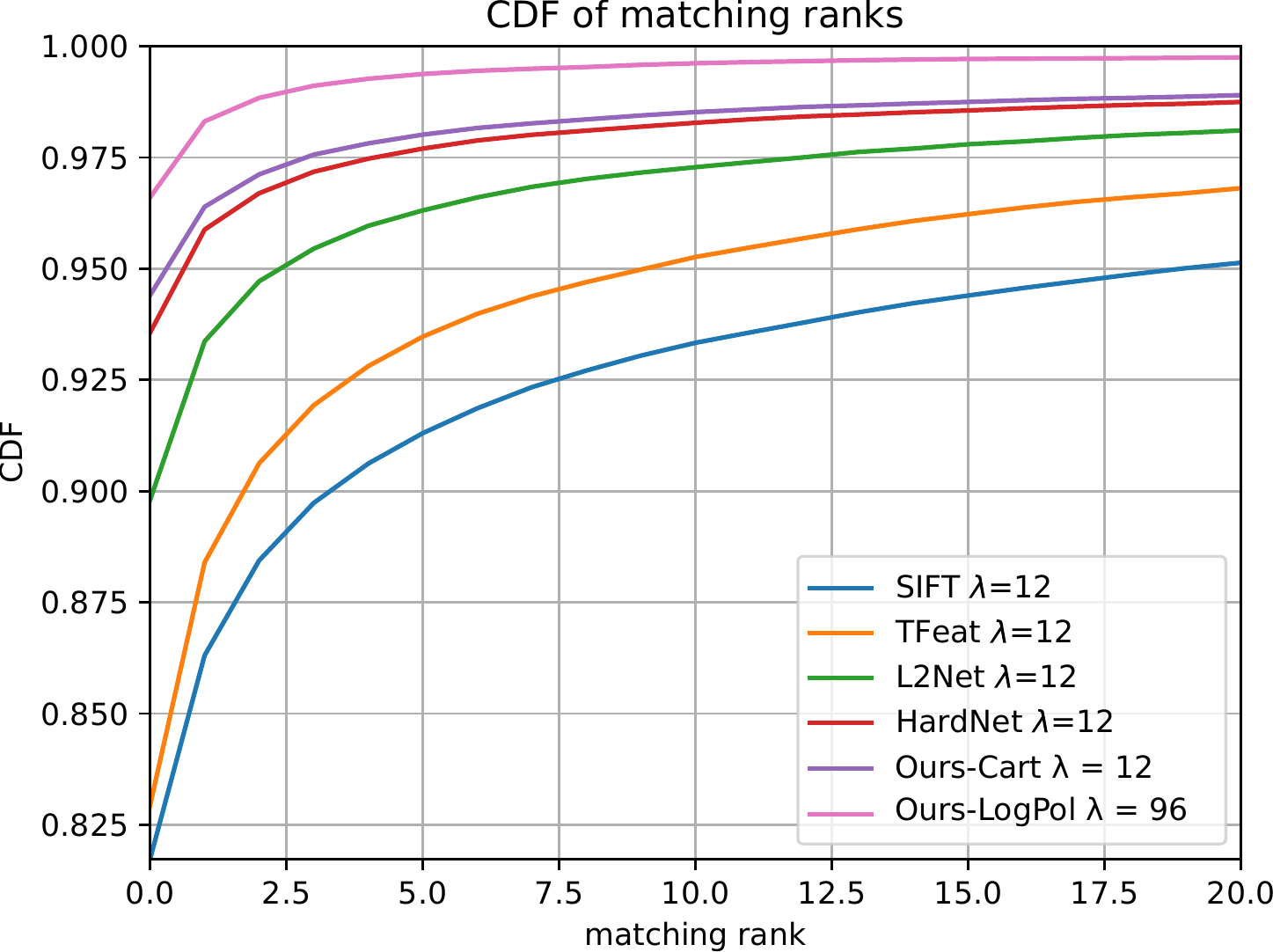}

\vspace{-3mm}
\caption{{\bf Patch retrieval on the new dataset.} We plot the cumulative distribution function of the rank in a patch retrieval scenario with a large number of distractors. Our models outperform all the baselines. Log-polar models ({\bf \color{Rhodamine}{pink}}) are significantly better than cartesian ones ({\bf \color{Plum}{purple}}) and baselines based on cartesian patches, such as HardNet ({\bf \color{Red}{red}}).}
\label{fig:ranking}
\vspace{-3mm}
\end{figure}

Next, we evaluate our performance in terms of patch retrieval. For every image pair in the test sequence, we extract SIFT keypoints on each image and establish ground truth correspondences using the procedure outlined in \refsec{dataset}. Matches with a difference of up to $25$ degrees in orientation are considered positives.
Typically, a large percentage of the image pixels are occluded, so that it is not possible to generate a large number of matches.
Instead, for every pair of images, we extract up to $N_m=500$ matches and then generate $N_d=3000$ distractors, defined as keypoints further than 3 pixels away from a keypoint.
The task is thus to find the ‘needle in the haystack’, where every keypoint has one positive match and $N_m + N_d - 1$ negatives.
We compute the distance between descriptors, extract the rank of each match, and accumulate it over all keypoints and images pairs.
The results are summarized in \fig{ranking}. Our models with log-polar patches obtain the best results, retrieving the correct match about 97\% of the time for our best model, for $\lambda=96$.
They are followed by our models with cartesian patches, and HardNet. Notice that contrary to the previous experiment, we evaluate 
on a realistic patch retrieval scenario with a large number of distractors, which indicates that our performance holds even when sampling keypoints densely, and that it does so regardless of $\lambda$.

\renewcommand{\arraystretch}{1.2}
\renewcommand{\tabcolsep}{2mm}
\begin{table}
\footnotesize
\begin{center}

\begin{tabular}{@{}lcccccc@{}}
	\toprule
	$\lambda$ & 12 & 16 & 32 & 64 & 96 & 128 \\
	\midrule
	Ours, Cart   & {\bf 0.72}  & 0.77 & 1.36 & 4.79 & 7.03 & 8.43 \\
	Ours, LogPol & 0.67 & 0.61 & 0.47 & 0.40 & {\bf 0.36} & {\bf 0.36} \\
	\bottomrule
\end{tabular}
\end{center}

\vspace{-3mm}
\caption{
{\bf FPR95 vs $\lambda$.} We evaluate models trained with differently-sized support regions. Performance increases with $\lambda$ for log-polar patches, but quickly degrades for cartesian ones.}
\label{tbl:results_lambda}
\end{table}

\subsection{Results on HPatches}
\label{sec:exp_hpatches}

The HPatches dataset~\cite{Balntas17} contains 116 sequences with 6 images each, with either viewpoint or illumination changes. As in~\cite{Brown10}, HPatches provides pre-extracted patches sampled at corresponding scales, which are not useful for our purposes. However, it also provides the original images and ground truth homographies.
We thus define the following protocol.
We use SIFT to find keypoints and determine correspondences among them using the ground truth homographies.
We consider sequences with viewpoint and illumination changes separately.
This provides us with 20733 correspondences for the illumination split and 22079 correspondences for the viewpoint split.
For every match, we compute the distance between a pair of corresponding descriptors and also to all the negatives in the dataset, and evaluate our models in terms of the rank-1 metric, \ie, the percentage of samples for which we can retrieve the correct match with rank 1. We show the results in \tbl{results_hpatches}. As expected, our log-polar models outperform most of the baselines%
, and perform better as $\lambda$ increases. For this experiment we use the models trained on our dataset, without fine-tuning.

\subsection{Results on AMOS patches}
\label{sec:exp_amos}

We also consider AMOS patches~\cite{Pultar19}, a dataset released recently featuring pairs of images captured by webcams and carefully curated in order to provide correspondences. We evaluate our method on the training split, which consists of 27 sequences, each with 50 images, and which also provides keypoints with scales and orientations for every image.
We use unique matching keypoint pairs across all images, obtaining a split of 13268 unique keypoint pairs. We use the same metric as
for HPatches,
and summarize the results in \tbl{results_amos}. As before, we do not re-train the models in any way. Again, our models outperform the state of the art and our results improve with the size of the support region, unlike for methods based on cartesian patches.

\renewcommand{\arraystretch}{1}
\renewcommand{\tabcolsep}{2mm}
\begin{table}
\small
\begin{center}
\begin{tabular}{@{}lcc@{}}
\toprule
Method                         & Viewpoint split & Illumination split \\
\midrule
SIFT, $\lambda=12$             & 0.740                  & 0.607                     \\
HardNet, $\lambda=12$          & 0.813                  & 0.707                     \\
GeoDesc, $\lambda=12$          & {\bf 0.879}            & 0.727                     \\
\midrule
Ours-Cart, $\lambda=12$        & 0.828                  & 0.722                     \\
Ours-Cart, $\lambda=16$        & 0.831                  & 0.732                       \\
Ours-Cart, $\lambda=32$        & 0.825                  & 0.736                     \\
Ours-Cart, $\lambda=64$        & 0.752                  & 0.666                     \\
Ours-Cart, $\lambda=96$        & 0.681                  & 0.616                       \\

\midrule
Ours, LogPol, $\lambda=12$      & 0.833                  & 0.729                     \\
Ours, LogPol, $\lambda=16$      & 0.838                  & 0.743                       \\
Ours, LogPol, $\lambda=32$      & 0.849                  & 0.764                     \\
Ours, LogPol, $\lambda=64$      & 0.849                  & 0.774                      \\
Ours, LogPol, $\lambda=96$      & 0.847                  & {\bf0.774}              \\
\bottomrule
\end{tabular}
\end{center}
\vspace{-3mm}
\caption{{\bf Results on HPatches.} Rank-1 performance on the viewpoint and illumination splits of the HPatches dataset. Our log-polar sampling approach performs better on average than all the baselines, and performance increases with $\lambda$, until it saturates.
}
\label{tbl:results_hpatches}
\end{table}

\renewcommand{\arraystretch}{1}
\renewcommand{\tabcolsep}{1.8mm}
\begin{table}
\small
\begin{center}
\begin{tabular}{@{}lcccccc@{}}
\toprule
Method & \multicolumn{5}{c}{Rank-1} \\
\midrule
$\lambda$ & 6 & 12 & 16 & 32 & 64 & 96 \\
\midrule
SIFT & 0.551 & 0.518 & 0.516 & 0.510 & 0.480 & 0.436 \\
GeoDesc & 0.434 & 0.396 & 0.389 & 0.416 & 0.438 & 0.417 \\
HardNet & 0.529 & 0.464 & 0.450 & 0.451 & 0.470 & 0.449 \\
Ours, Cart & 0.554 & 0.507 & 0.530 & 0.549 & 0.524 & 0.481 \\
Ours, LogPol & 0.607 & 0.604 & 0.625 & 0.641 & 0.648 & {\bf 0.651} \\
\bottomrule
\end{tabular}
\end{center}
\vspace{-4mm}
\caption{{\bf Results on AMOS patches.} Rank-1 performance on the AMOS patches dataset. We noticed that for this dataset, extracting descriptors with smaller patches produces better results for most baselines, so we also consider $\lambda=6$.
Our models trained on log-polar patches outperform the state of the art, and performance increases with $\lambda$.
}
\vspace{-2mm}
\label{tbl:results_amos}
\end{table}

\et{
\begin{table}[t]
\begin{center}
\renewcommand{\arraystretch}{0.8}
\renewcommand{\tabcolsep}{1.3mm}
\footnotesize
\begin{tabularx}{\linewidth}{llcccc}
  \toprule
  \multirow{2}{*}{Type} & \multirow{2}{*}{Method} & \multicolumn{2}{c}{Stereo task} & \multicolumn{2}{c}{Multi-view task} \\
  & & mAP$^{15^o}$ & Rank$^\dag$ & mAP$^{15^o}$ & Rank$^\dag$ \\
  \midrule
  \multirow{6}{*}{DoG} & SIFT (IJCV'04)            & 0.0277 & 9 & 0.4146 & 8 \\
  & TFeat (BMVC'16)           & 0.0357 & 8 & 0.4643 & 7 \\
  & L2-Net (CVPR'17)          & 0.0400 & 6 & 0.5087 & 5 \\
  & HardNet (NIPS'17)         & 0.0425 & 4 & {\bf \color{red}{0.5481}} & 1 \\
  & GeoDesc (ECCV'18)         & 0.0368 & 7 & 0.5298 & 4 \\
  & ContextDesc (CVPR'19)     & 0.0439 & 3 & 0.5399 & 3 \\
  \midrule
  \multirow{2}{*}{e2e} & SuperPoint (CVPR'18)      & 0.0415 & 5 & 0.4778 & 6 \\
  & D2-Net (CVPR'19)          & {\bf \color{red}{0.0490}} & 1 & 0.3967 & 9 \\
  \midrule
  \multirow{4}{*}{\shortstack[l]{Ours\\ (DoG)}}& Ours-Cartesian, $\lambda=16$ & 0.0405 & --- &  0.5208& --- \\
  & Ours-LogPol, $\lambda=32$ & 0.0420 & --- & 0.5389 & --- \\
  & Ours-LogPol, $\lambda=64$ & 0.0432 & --- & 0.5396 & --- \\
  & Ours-LogPol, $\lambda=96$ & {\bf \color{ForestGreen}{0.0448}} & 2 & {\bf \color{ForestGreen}{0.5427}} & 2 \\
  \bottomrule
\end{tabularx}
\vspace{-1.5em}
\end{center}
\caption{{\bf PhotoTourism challenge.} Mean average precision in pose estimation with an error threshold of 15$^o$.
Top method ($^\dag$among comparable submissions) in {\bf \color{red}{red}}, runner-up in {\bf \color{ForestGreen}{green}}. We rank 2nd on both tracks, and 1st on average.}
\label{tbl:results_imw}
\vspace{-1em}
\end{table}

}

\subsection{Results on the Phototourism Challenge}
\label{sec:exp_challenge}

Patch matching performance does not always translate to upstream applications, as evidenced by~\cite{Yi16b,Schonberger17}. We thus also evaluate our method on the public Phototourism Image Matching challenge~\cite{imwchallenge2019}.
This benchmark features two tracks: stereo and multi-view matching, and evaluates local features in terms of the quality of the {\em reconstructed poses}. Features are submitted to the organizers, who compute the results. We provide them in \tbl{results_imw}, including comparable baselines (up to 8k features per image, matched by brute-force nearest-neighbour) extracted from the public leaderboards.
Our method ranks second on both tracks, and first in terms of average rank.
Note that our observations from~\secref{exp_patches} carry over -- models trained on log-polar patches improve with patch size, and outperform cartesian models.

\section{Conclusions and Future Work}
\label{sec:conclusions}

We have introduced a novel approach to learn local descriptors that goes beyond the current paradigm,
which relies on image measurements sampled in cartesian space. %
We show that we can learn richer and more scale-invariant representations by coupling log-polar sampling with state-of-the-art deep networks. This allows us to match local descriptors across a wider range of scales, virtually for free.

Our approach could be used to learn invariance to arbitrary scale changes. This can be, however, counterproductive when used alongside SIFT, as the majority of its detections are accurate enough.
Instead, we intend to bypass scale detection and learn end-to-end pipelines as in~\cite{Yi16b,Ono18}.

{\small
\bibliographystyle{ieee_fullname}
\bibliography{short,vision,learning,sites}
}

\onecolumn
\setcounter{page}{1}
\section{Beyond Cartesian Representations for Local Descriptors: Supplementary Material}

\subsection{Regarding the Dataset}
\label{sec:app_data}

In order to train scale-invariant models with real data relevant to wide-baseline stereo, it was necessary to collect training data. For this we rely on public collections of photo-tourism images in the Yahoo Flickr Creative Commons 100M (YFCC) dataset. We use COLMAP, a Structure from Motion (SfM) framework, to obtain 3D reconstructions. COLMAP provides us with sparse point clouds and depth maps for each image. We clean up the depth maps following the procedure outlined in the paper and use them, along with the ground truth camera poses, to project keypoints between corresponding images.

We sample pairs of images with a visibility check in order to guarantee that a minimum number of keypoints can be extracted and matched across both views. Specifically, we retrieve the SfM keypoints in common over both views, extract their bounding box, and reject the image pair if it is smaller than a given threshold (we use 0.5) for either image.

\begin{figure}[H]
\centering
\renewcommand{\imsize}{1.6cm}
\newcommand{\imscl}{0.01}
\includegraphics[height=\imsize]{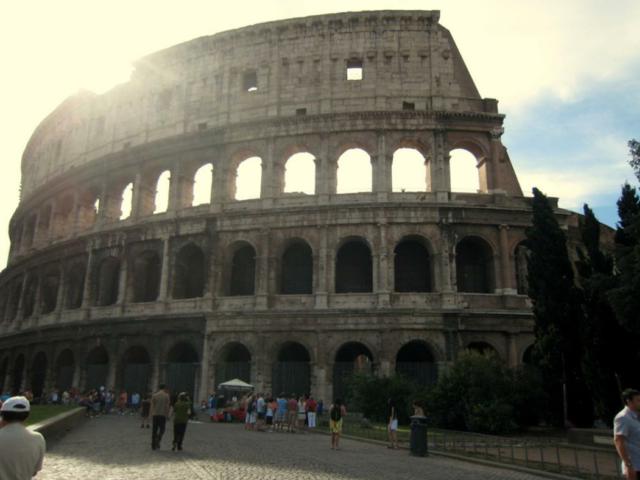}\hspace{\hspacer}%
\includegraphics[height=\imsize]{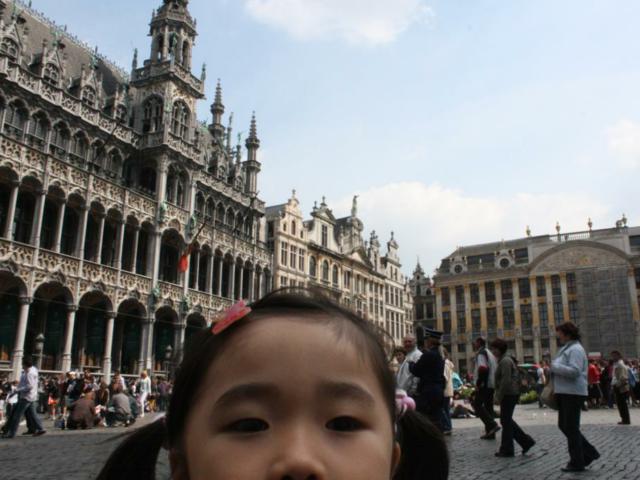}\hspace{\hspacer}%
\includegraphics[height=\imsize]{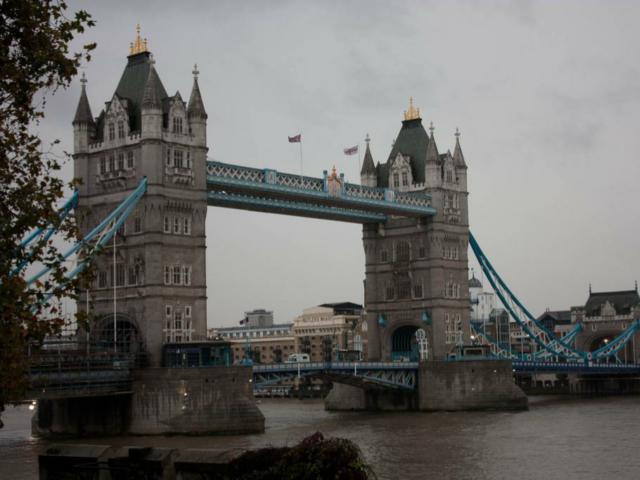}\hspace{\hspacer}%
\includegraphics[height=\imsize]{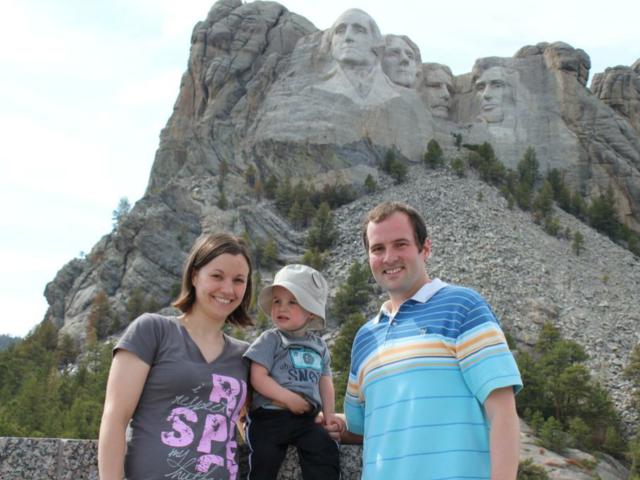}\hspace{\hspacer}%
\includegraphics[height=\imsize]{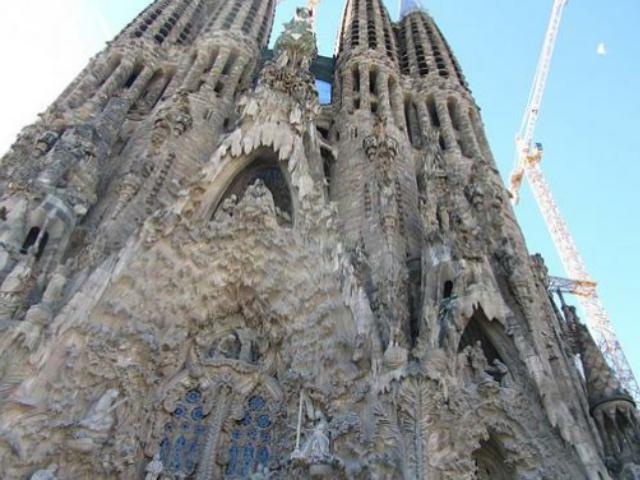}\hspace{\hspacer}%
\includegraphics[height=\imsize]{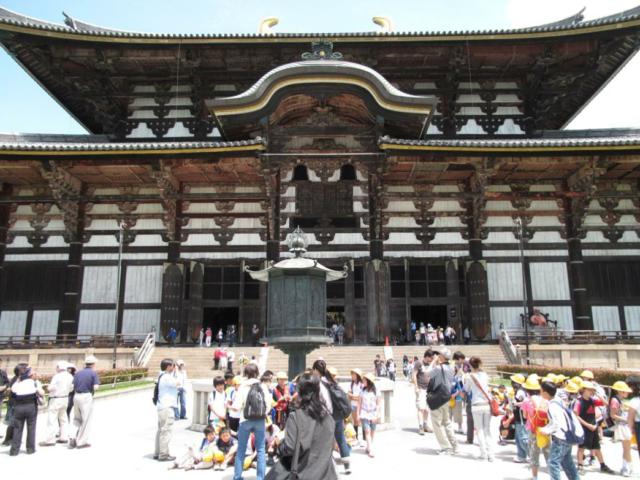}\hspace{\hspacer}%
\includegraphics[height=\imsize]{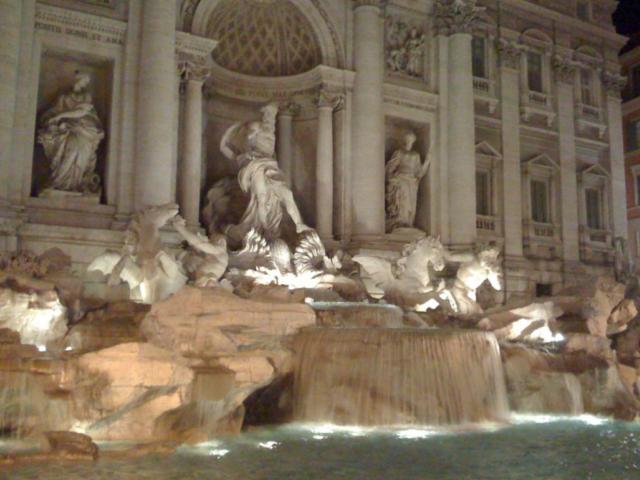}\hspace{\hspacer}%
\includegraphics[height=\imsize]{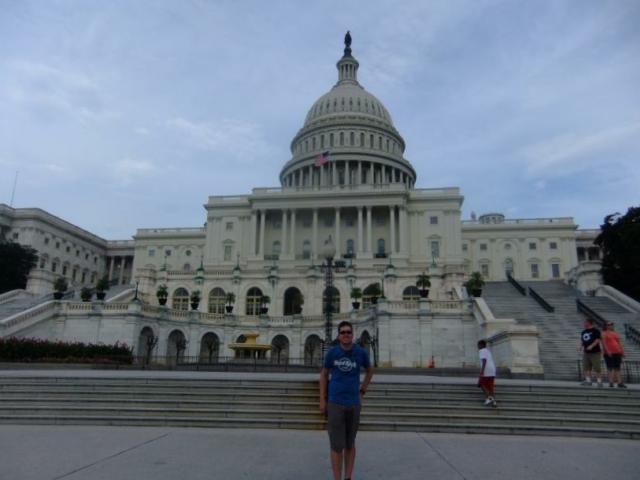}\\
\vspace{\vspacer}%
\includegraphics[height=\imsize]{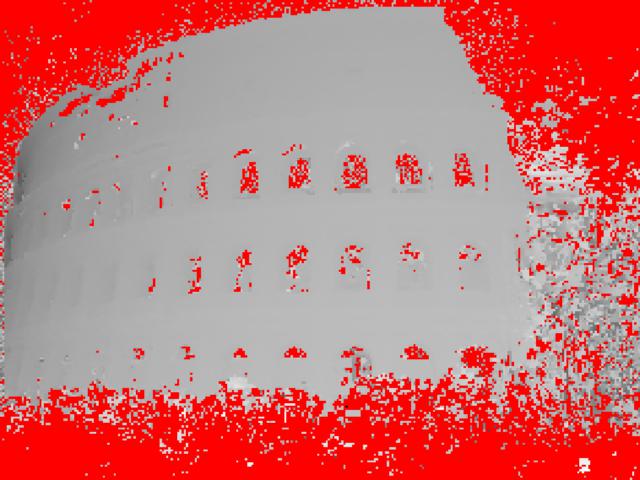}\hspace{\hspacer}%
\includegraphics[height=\imsize]{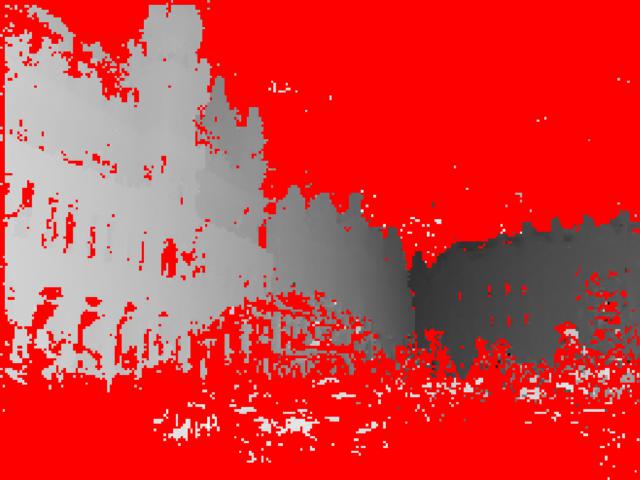}\hspace{\hspacer}%
\includegraphics[height=\imsize]{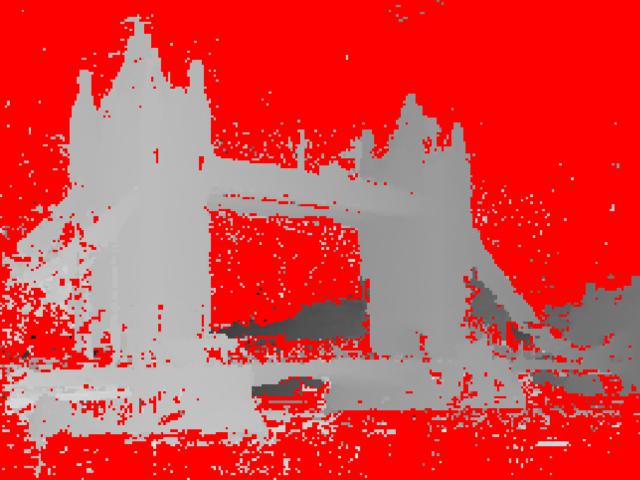}\hspace{\hspacer}%
\includegraphics[height=\imsize]{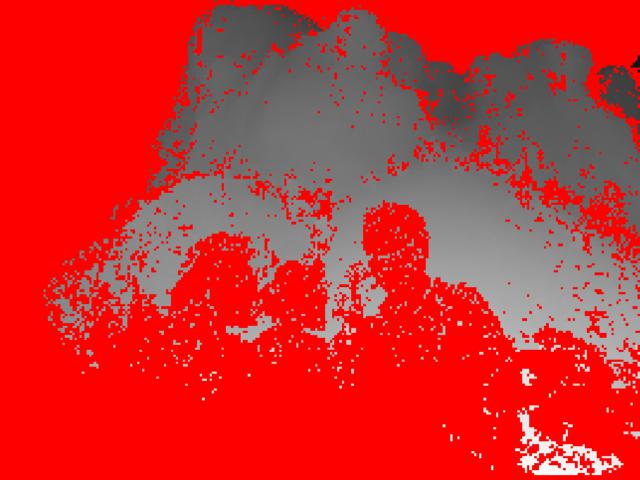}\hspace{\hspacer}%
\includegraphics[height=\imsize]{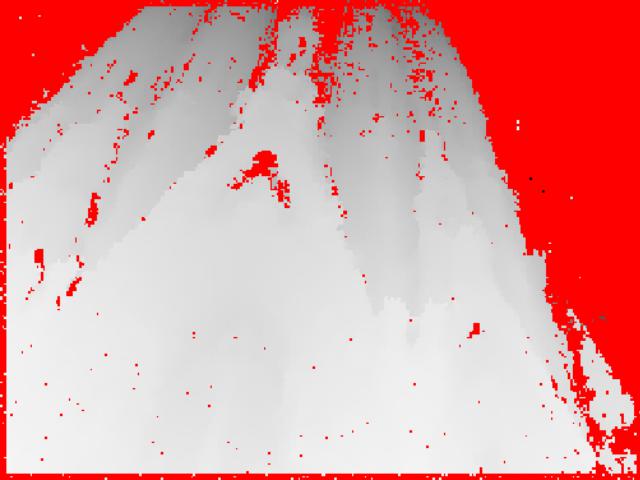}\hspace{\hspacer}%
\includegraphics[height=\imsize]{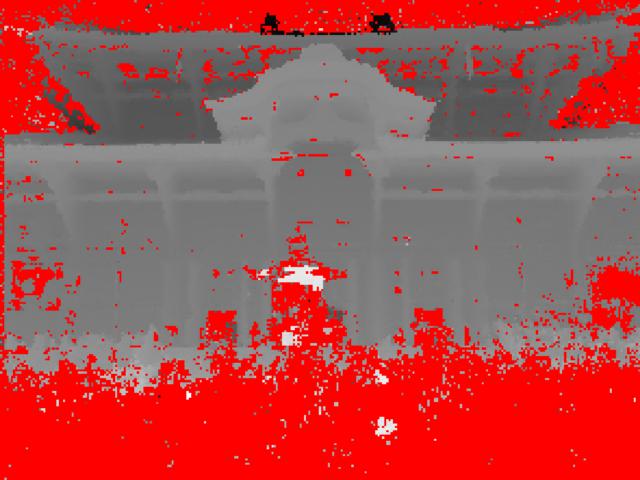}\hspace{\hspacer}%
\includegraphics[height=\imsize]{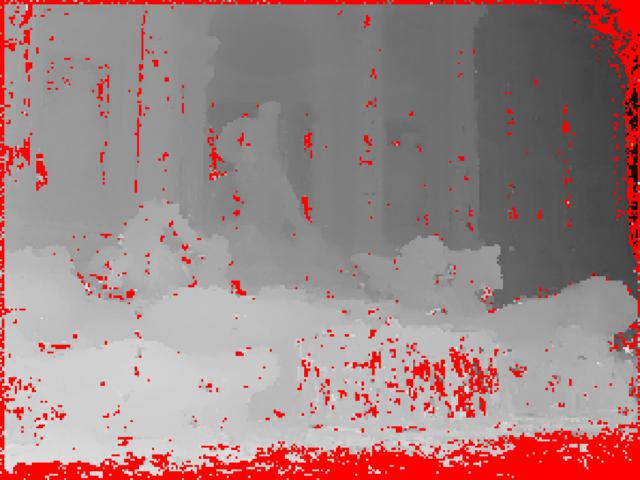}\hspace{\hspacer}%
\includegraphics[height=\imsize]{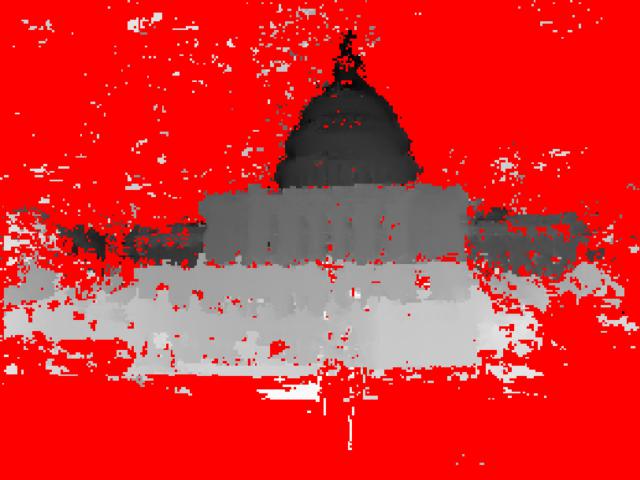}\\
\caption{{\bf Dataset samples.} We show the original images and their corresponding depth maps, estimated by COLMAP and post-processed by us as explained in \refsec{dataset_data}. The depth maps are color-coded by depth, in grayscale, with red indicating occlusions and regions for which depth estimates are not available. Notice how despite some noise and occluded areas, the depth estimates are good enough to extract training data.}
\label{fig:dataset_samples}
\end{figure}

We use 11 sequences for training and validation and 9 for testing. We list their details in \tbl{data_sequences}, and give some examples in \fig{dataset_samples}. This data will be made publicly available along with code and pre-trained models.

\renewcommand{\arraystretch}{1}
\renewcommand{\tabcolsep}{1mm}

\begin{table}[h]
\begin{minipage}{0.5\linewidth}
\centering
\begin{tabular}{@{}lc@{}}
\toprule
Sequence name & Num. images \\
\midrule
brandenburg\_gate & 1363 \\
buckingham\_palace & 1676 \\
colosseum\_exterior & 2063 \\
grand\_place\_brussels & 1083 \\
notre\_dame\_front\_facade & 3765 \\
palace\_of\_westminster & 983 \\
pantheon\_exterior & 1401 \\
sacre\_coeur & 1179 \\
st\_peters\_square & 2504 \\
taj\_mahal & 1312 \\
temple\_nara\_japan & 904 \\
\midrule
Total & 18233  \\
\bottomrule
\end{tabular}
\end{minipage}
\begin{minipage}{0.5\linewidth}
\begin{tabular}{@{}lc@{}}
\toprule
Sequence name & Num. images \\
\midrule
british\_museum & 660 \\
florence\_cathedral\_side & 108 \\
lincoln\_memorial\_statue & 850 \\
milan\_cathedral & 124 \\
mount\_rushmore & 138 \\
reichstag & 75 \\
sagrada\_familia & 401 \\
st\_pauls\_cathedral & 615 \\
united\_states\_capitol & 258 \\
\midrule
Total & 4107 \\
\bottomrule
\end{tabular}
\end{minipage}
\caption{{\bf Dataset details.} Left: training sequences. Right: Test sequences.}
\label{tbl:data_sequences}
\end{table}

\end{document}